\documentclass[journal]{IEEEtran}
\usepackage{amsmath,graphicx,cite,epsfig}
\usepackage{multirow}
\usepackage{booktabs}
\usepackage{float}
\usepackage{subfigure}
\usepackage{caption}
\usepackage{threeparttable}
\usepackage{diagbox}
\usepackage{url}
\usepackage{bm}
\usepackage{algorithm}
\usepackage{algorithmic}
\usepackage{bm}
\usepackage{amsfonts}

\usepackage[table,xcdraw]{xcolor}
\ifCLASSINFOpdf

\else

\fi

\begin{document}
%
\title{Source-free Unsupervised Domain Adaptation for Blind Image Quality Assessment}
%
%
%

\author{Jianzhao Liu, Xin Li, Shukun An  and  Zhibo Chen,~\IEEEmembership{Senior~Member,~IEEE}
\thanks{J. Liu,  X. Li, S. An and Z. Chen are with the CAS Key Laboratory of Technology in Geo-Spatial Information Processing and Application System, University of Science and Technology of China, Hefei 230027, China (e-mail: jianzhao@mail.ustc.edu.cn;  lixin666@mail.ustc.edu.cn; ask@mail.ustc.edu.cn;  chenzhibo@ustc.edu.cn).}}
\maketitle

\begin{abstract}
Existing learning-based methods for blind image quality assessment (BIQA) are heavily dependent on large amounts of annotated training data, and usually suffer from a severe performance degradation when encountering the domain/distribution shift problem. Thanks to the development of unsupervised domain adaptation (UDA), some works attempt to transfer the knowledge from a label-sufficient source domain to a label-free target domain under domain shift with UDA. However, it requires the coexistence of source and target data, which might be impractical for source data due to the privacy or storage issues.
 In this paper, we take the first step towards the  source-free unsupervised domain adaptation (SFUDA)  in a simple yet efficient manner for BIQA to tackle the domain shift without access to the source data. 
 Specifically, we cast the quality assessment task as a rating distribution prediction problem. Based on the intrinsic properties of BIQA, we present a group of well-designed self-supervised objectives to guide the adaptation of the BN affine parameters towards the target domain. Among them, minimizing the prediction entropy and maximizing the batch prediction diversity aim to encourage more confident results while avoiding the trivial solution. 
 Besides, based on the observation that the IQA rating distribution of single image follows the Gaussian distribution,
 we apply Gaussian regularization to the predicted rating distribution to make it more consistent with the nature of human scoring. Extensive experimental results under cross-domain scenarios demonstrated the effectiveness of our proposed method to mitigate the domain shift.  
\end{abstract}




\maketitle

\section{Introduction}

\begin{figure}
\centering
\includegraphics[width=0.95\linewidth]{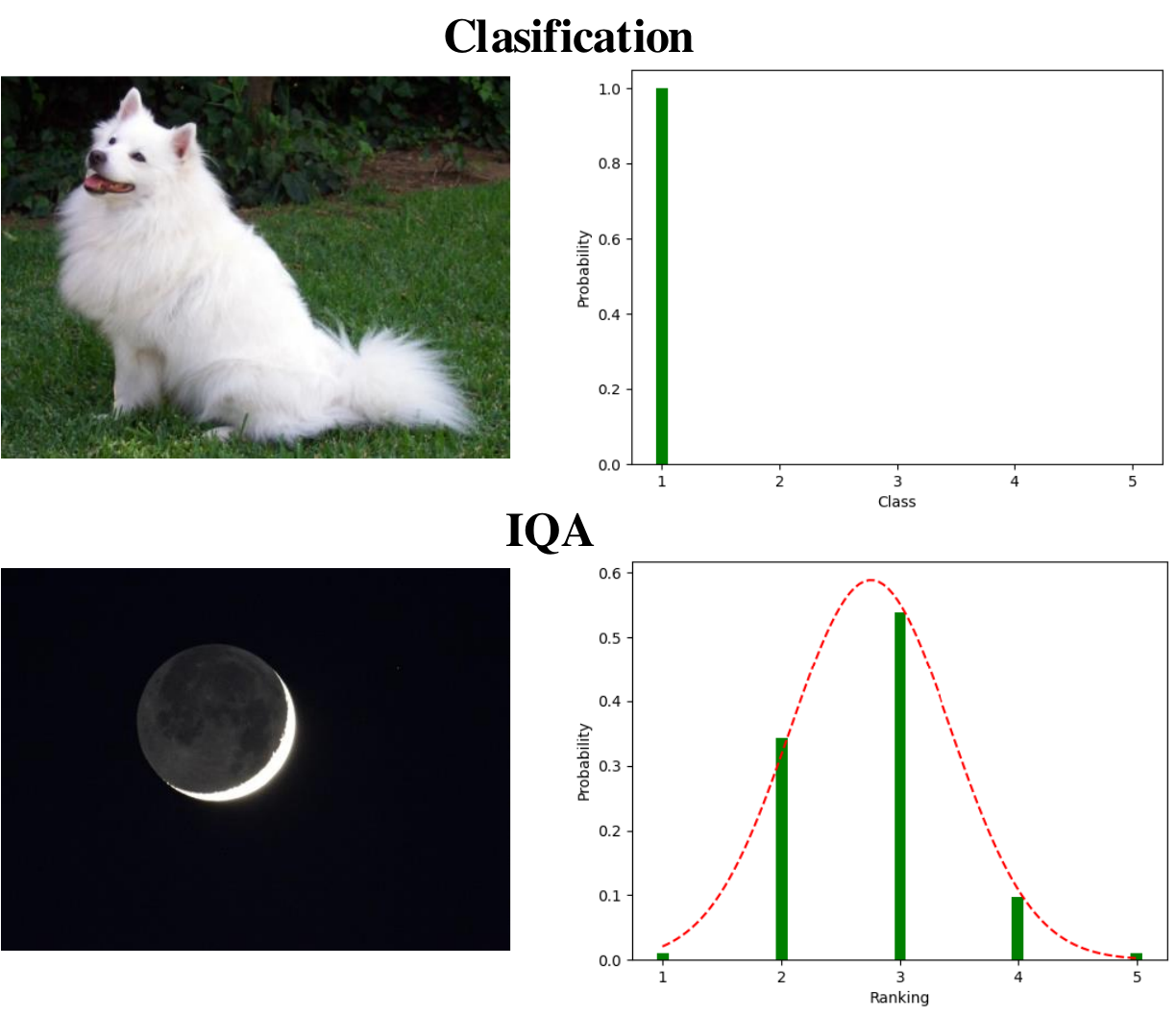}
\caption{Illustration of the difference regarding the label between classification task and IQA.}
\label{fig:class_vs_iqa}
\vspace{-2mm}
\end{figure}

Blind image quality assessment (BIQA) aims to automatically predict the  perceptual image quality without any information of reference images. Recent years have witnessed the impressive success of deep learning in BIQA area. However, learning-based methods rely heavily on abundant annotated data. For BIQA whose annotation process is laborious and time-consuming, it's hard to obtain large amount of data, leading to the poor generalization under the cross-domain/cross-dataset scenario. To alleviate the dependence of annotation, some works \cite{uncertaintyDA,rankDA} utilize the unsupervised domain adaptation (UDA) technique to transfer the learned knowledge from label-sufficient source domain to label-free target domain in IQA area. UCDA \cite{uncertaintyDA} adopts two-stage adaptation method, where in the first stage a coarse adaptation is applied between the source and target domain, and in the second stage a fine-level adaptation is applied between the confident and uncertain subdomains of target domain. Inspired by the transferability of pairwise quality relationship, RankDA~\cite{rankDA} adopts maximum mean discrepancy (MMD) loss to align the ranking features extracted from source and target domains. However, existing UDA methods for BIQA are required to access the source data when learning to adapt the model, which is impractical when the source data is not available during training (\textit{e.g.}, those from surveillance cameras or hospitals). In this paper, we address an interesting but challenging UDA scenario in IQA where only a trained source model and the unlabeled target data are provided.

A natural idea is to directly adopt the source-free unsupervised domain adaptation (SFUDA) methods invented in high-level classification tasks. For example, SHOT \cite{SHOT} fine-tunes the feature encoder by information maximization loss and self-supervised pseudo-labeling. Unfortunately, without taking into account the essential difference between classification and IQA, these methods will lead to inferior performance when applied directly to the IQA task. Conventional IQA methods only regress the scalar image quality score, while in this paper we cast the quality assessment task as a rating distribution prediction problem~\cite{NIMA,uncertaintyDA,zeng2017probabilistic} with a novel probabilistic modeling method, in order to bridge the gap between the classification and IQA. In spite of this, there still exist large differences that hamper the performance of applying high-level SFUDA methods to IQA.

\begin{figure*}[h!]
 \centering
 \subfigure[MOS distributions of KonIQ-10K]{
  \label{fig:koniq_allmos}
  \includegraphics[width = .22\textwidth]{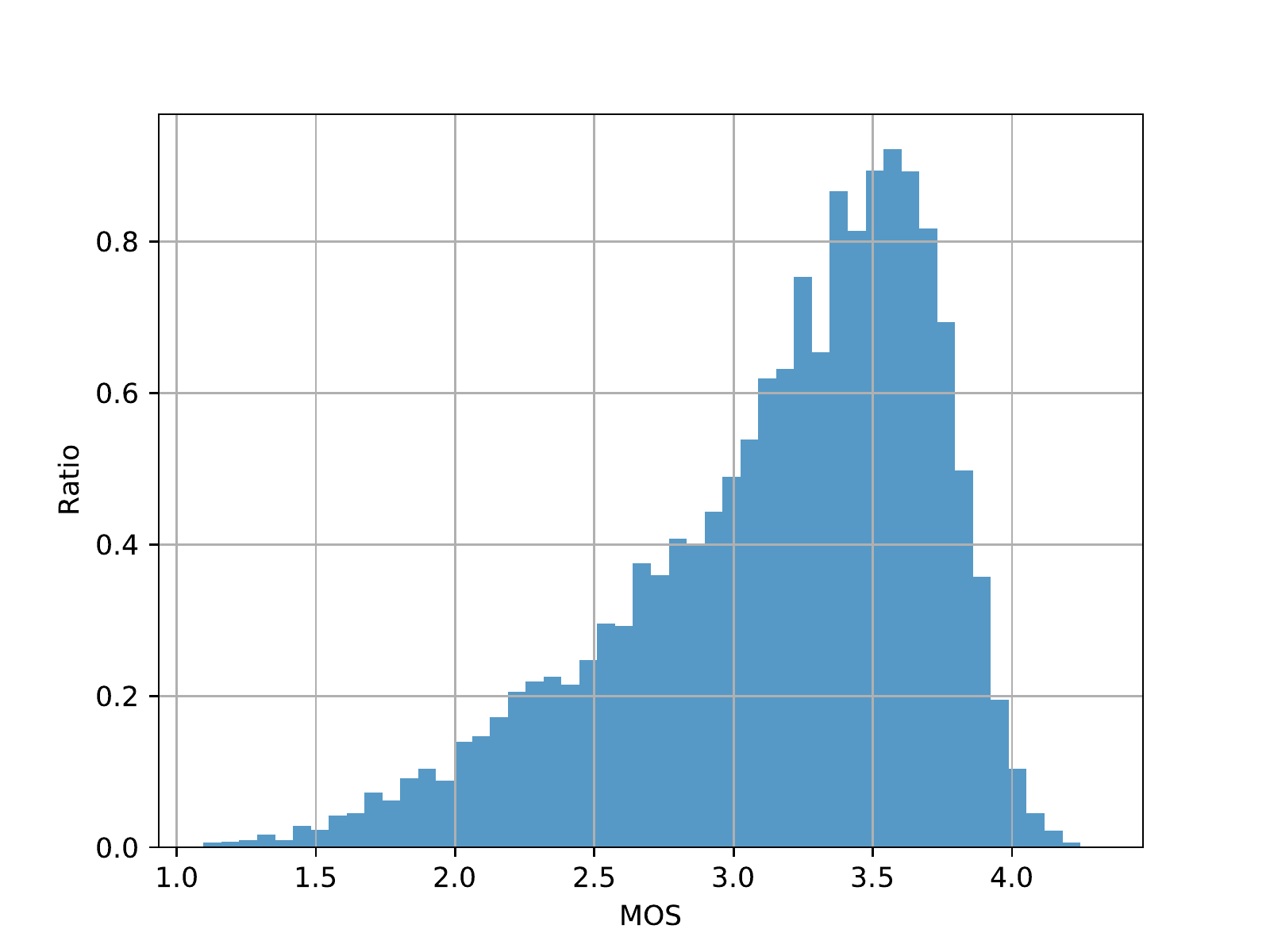}}
 \subfigure[$1.0<=MOS<1.5$]{
  \label{fig:1-1.5}
  \includegraphics[width = .22\textwidth]{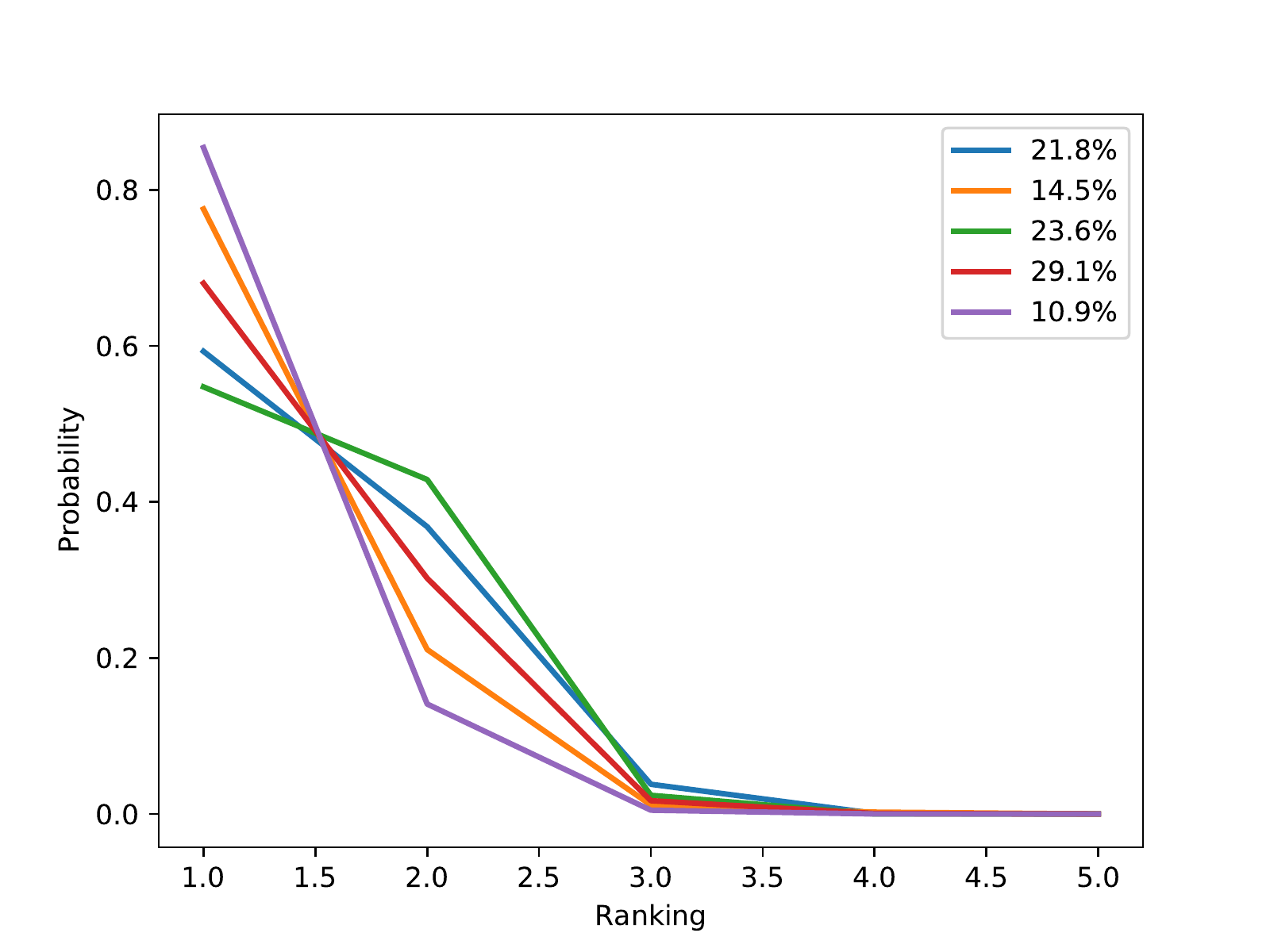}}
\subfigure[$1.5<=MOS<2.0$]{
\label{fig:1.5-2.0}
\includegraphics[width = .22\textwidth]{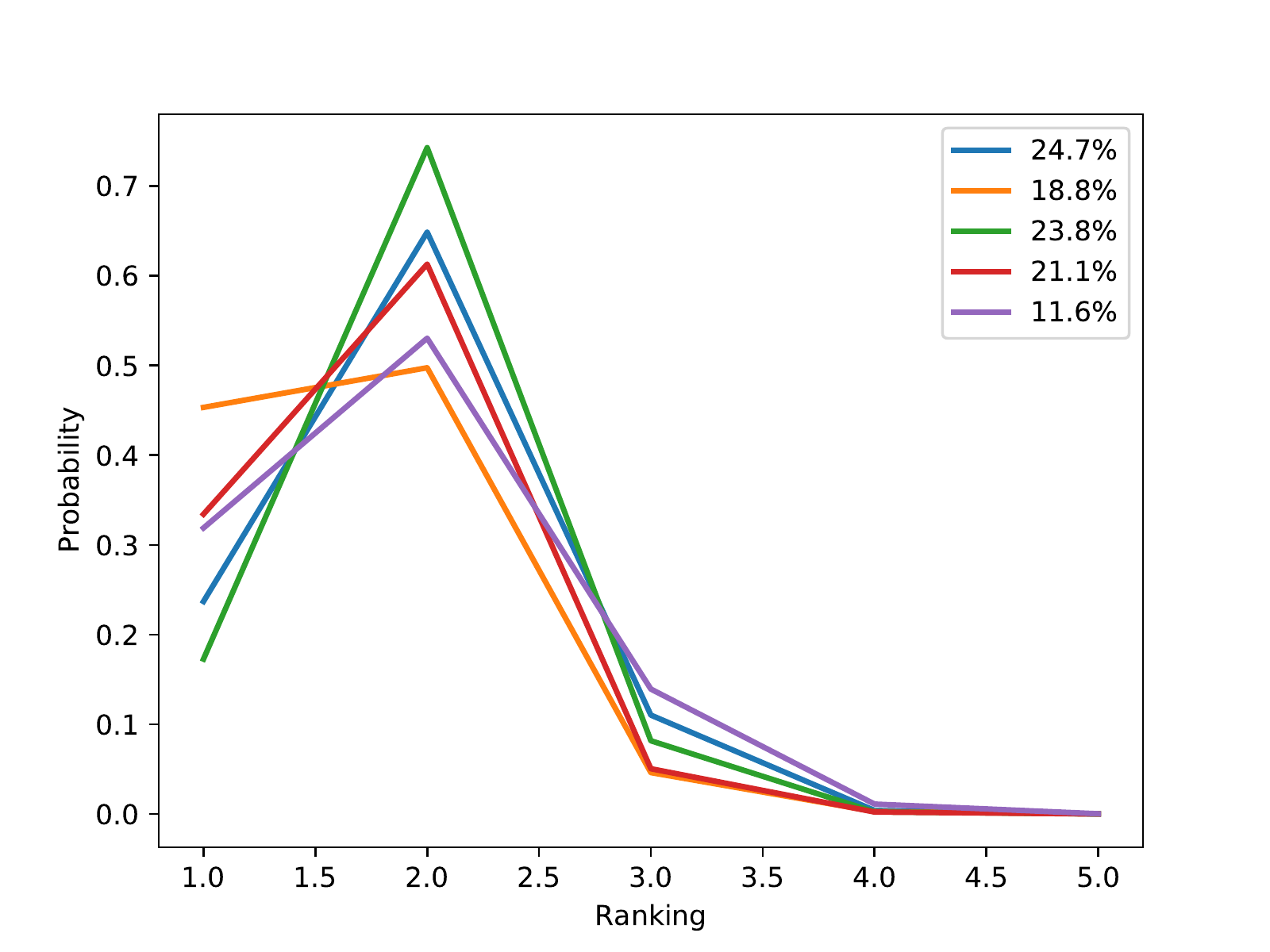}}
\subfigure[$2.0<=MOS<3.5$]{
\label{fig:2.0-2.5}
\includegraphics[width = .22\textwidth]{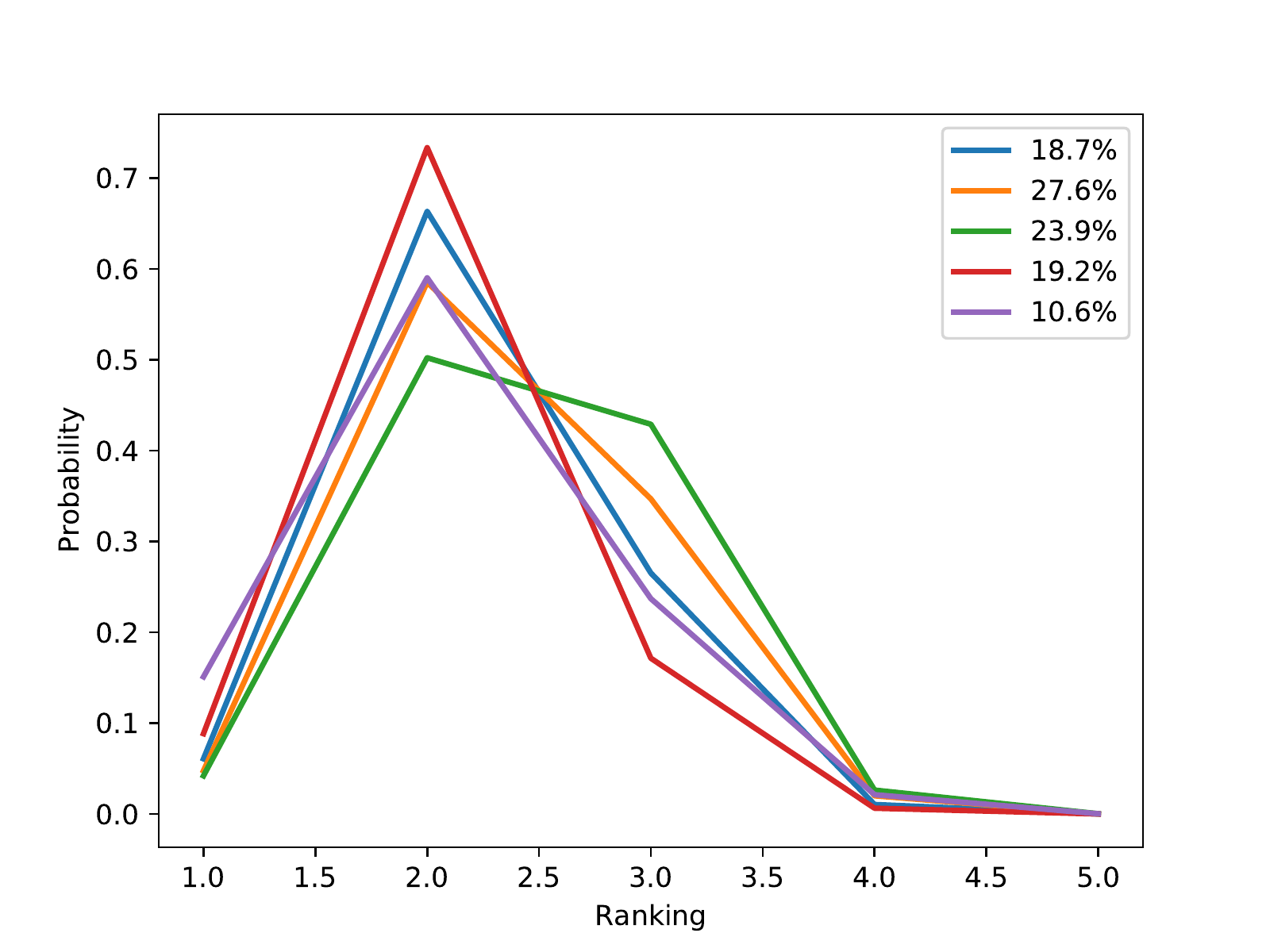}}

\subfigure[$2.5<=MOS<3.0$]{
  \includegraphics[width = .22\textwidth]{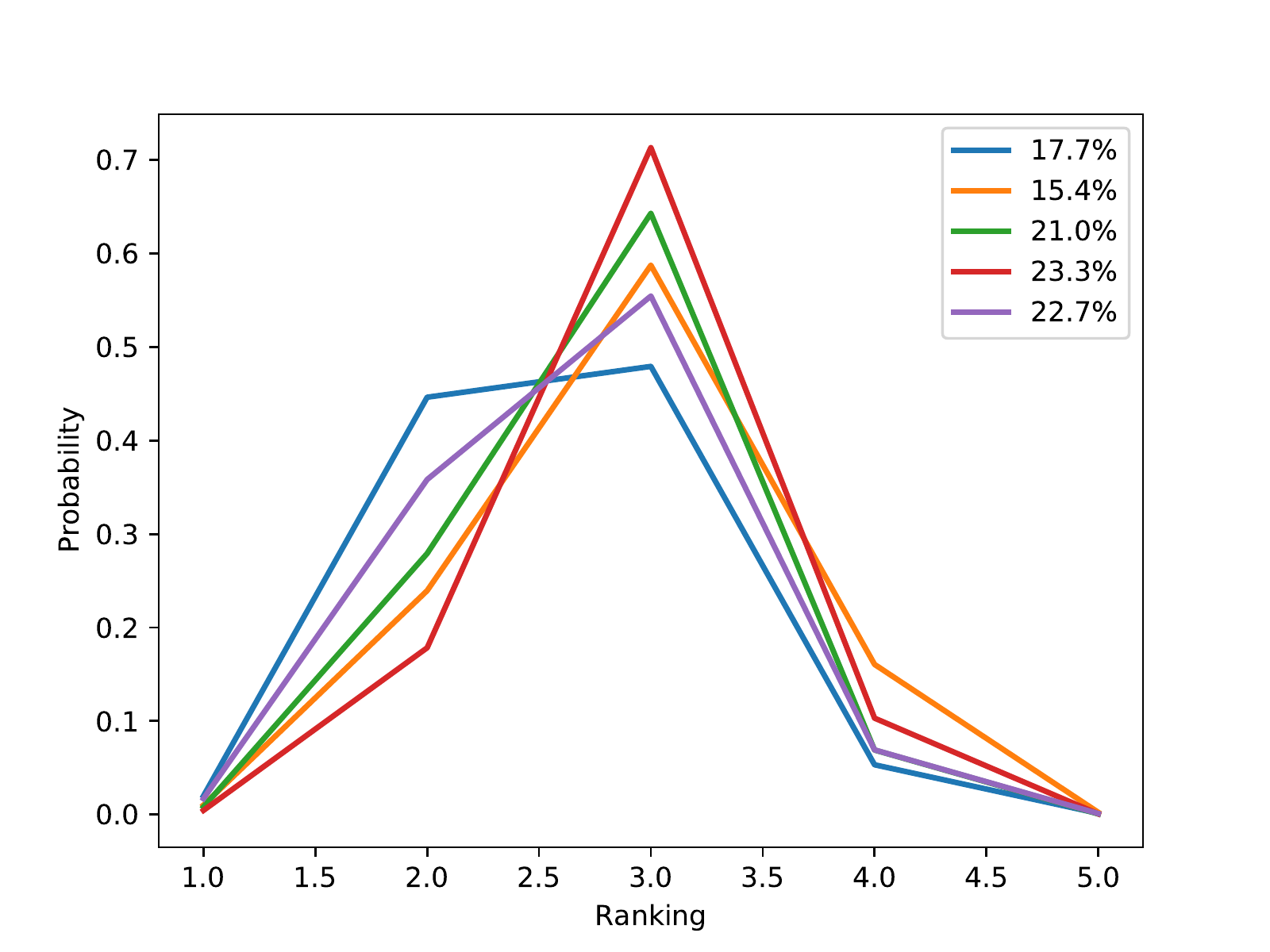}}
 \subfigure[$3.0<=MOS<3.5$]{
  \label{fig:1-1.5}
  \includegraphics[width = .22\textwidth]{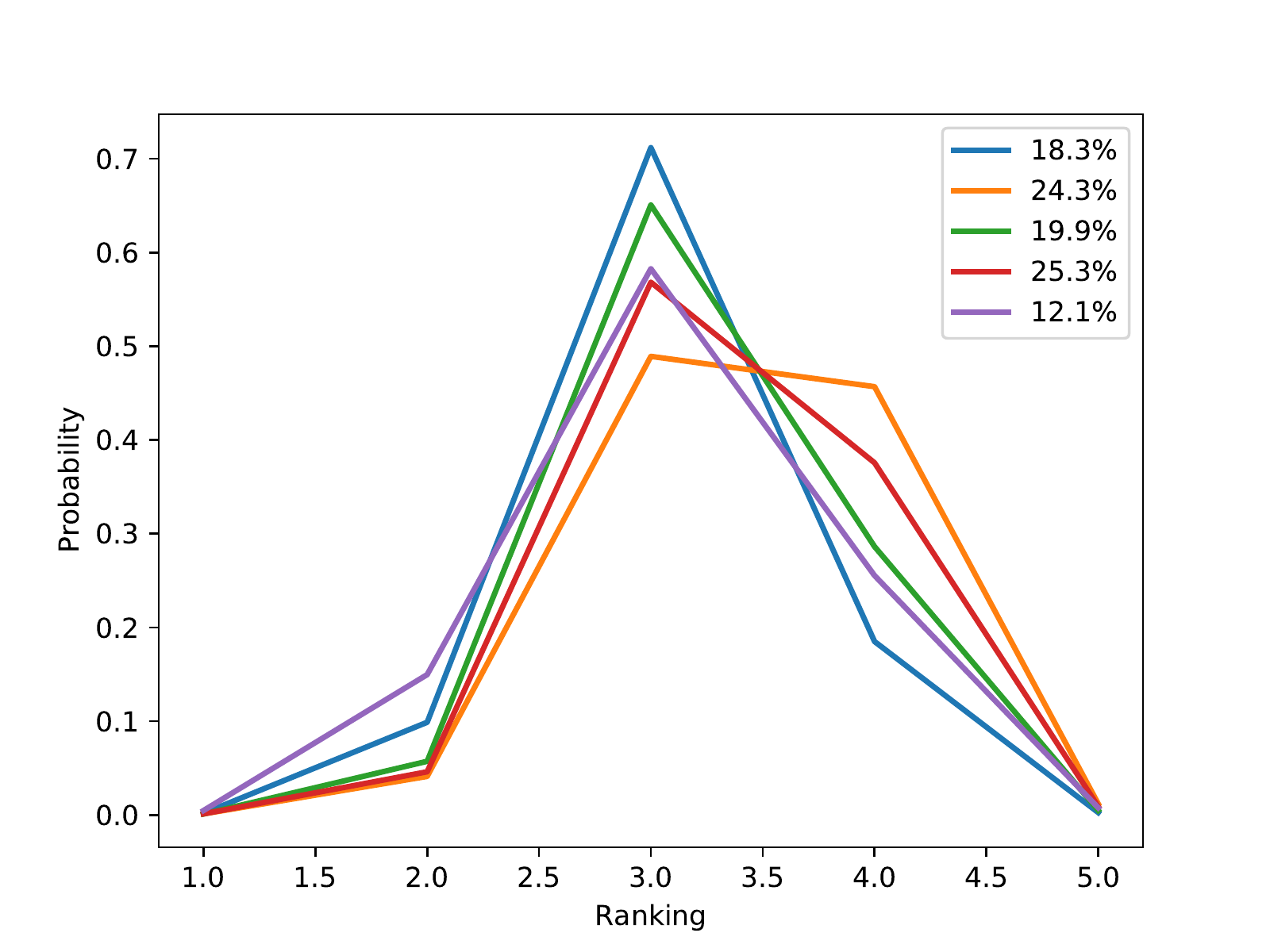}}
\subfigure[$3.5<=MOS<4.0$]{
\label{fig:1.5-2.0}
\includegraphics[width = .22\textwidth]{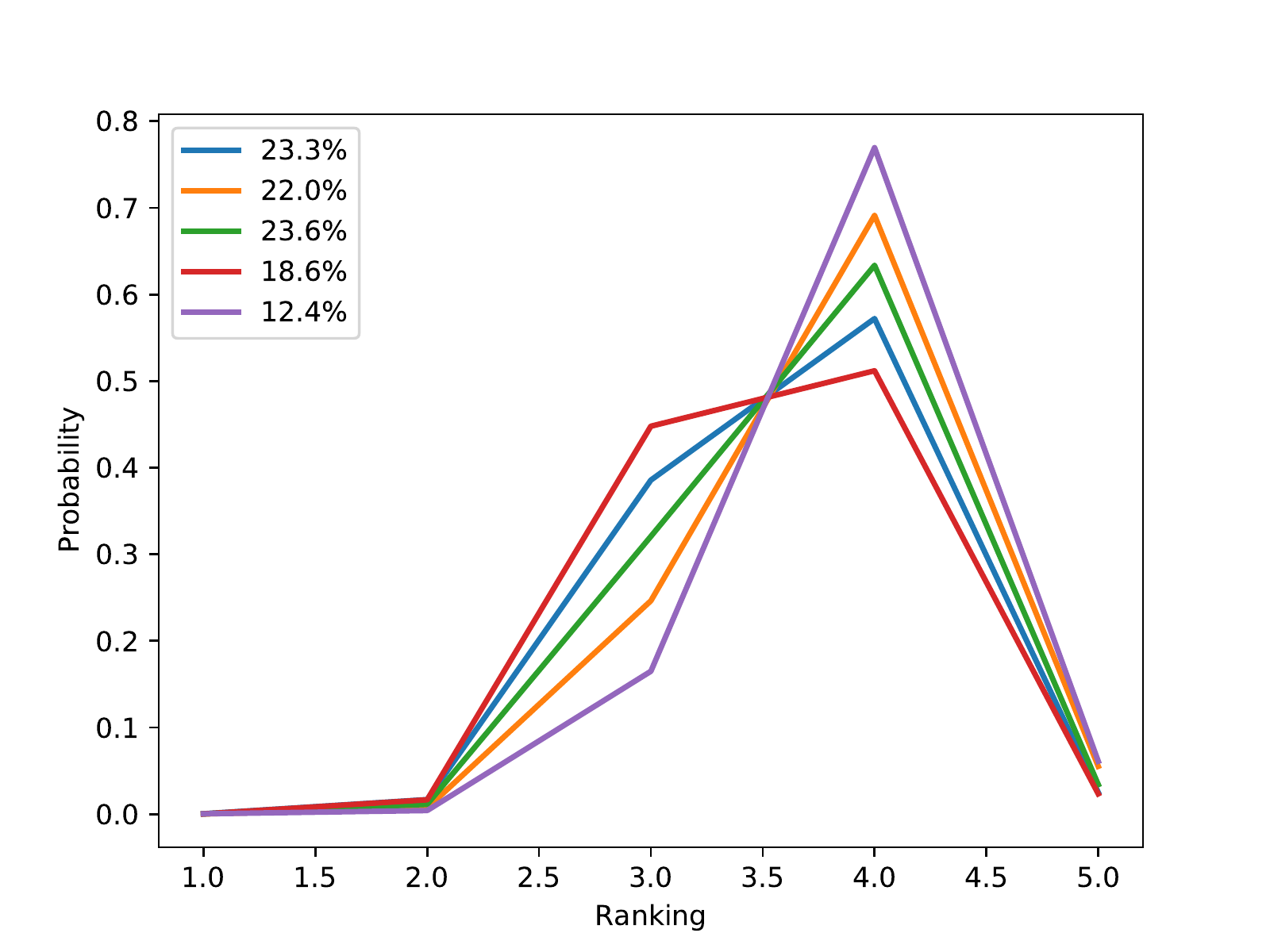}}
\subfigure[$4.0<=MOS<4.5$]{
\label{fig:2.0-2.5}
\includegraphics[width = .22\textwidth]{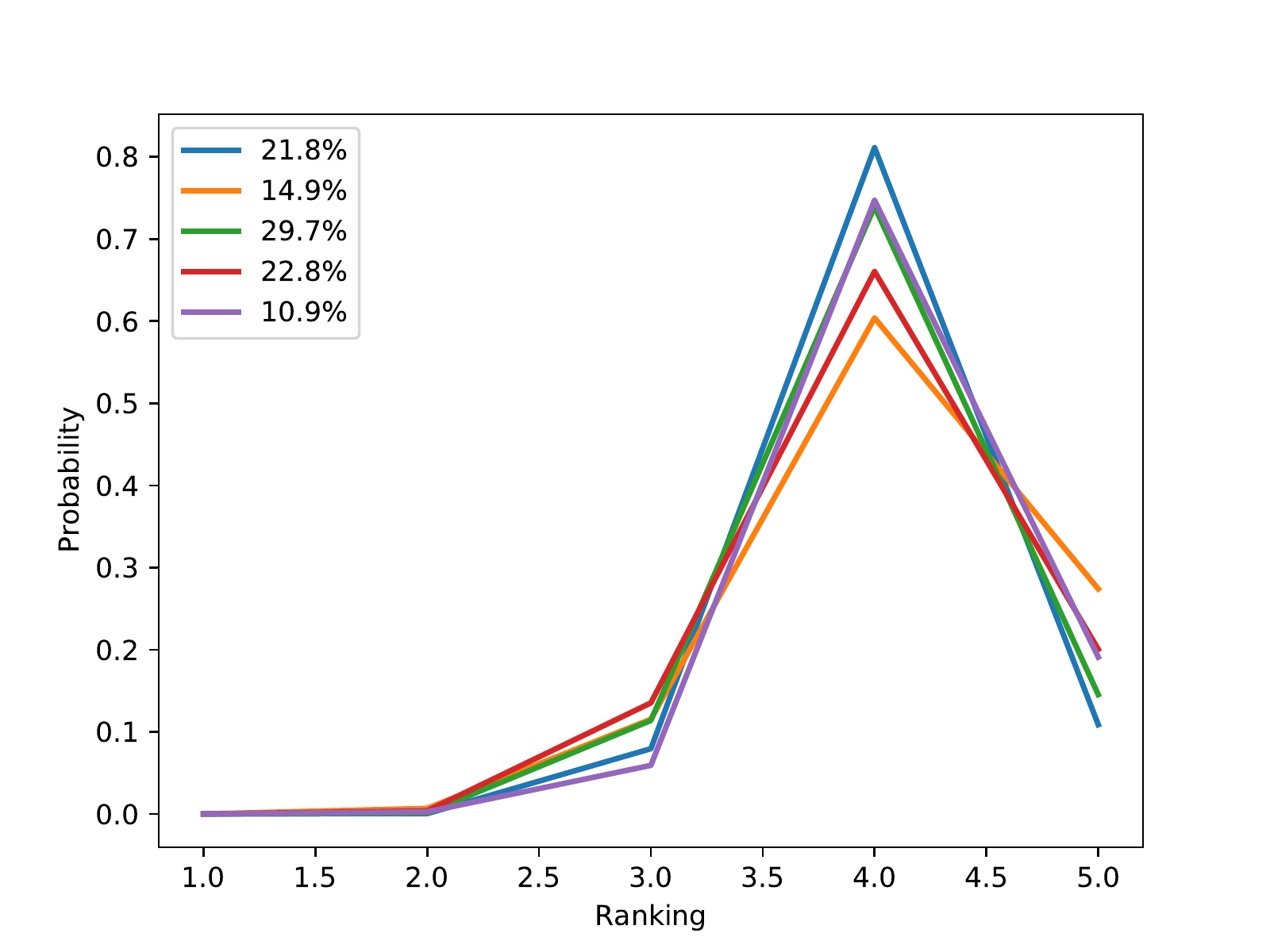}}
\caption{Clusters of distributions for images with different MOS ranges. The legend of each plot shows the percentage of these images associated with each cluster. The score distributions tend to be Gaussian. }
\label{fig:clusters}
\end{figure*}

\begin{table*}[]
\centering
\caption{Goodness-of-Fit(GoF) measured by RMSE between distributions we used to model the score distributions of KonIQ-10K. The percentage in the bracket denotes the percentage of these images whose MOSs are within the corresponding range. }
	\begin{tabular}{|c|c|c|c|c|c|c|c|c|}
		\hline
		\diagbox{Model}{RMSE}{MOS range}
		 & \begin{tabular}[c]{@{}c@{}}1.0-1.5\\ (0.546\%)\end{tabular} &\begin{tabular}[c]{@{}c@{}}1.5-2.0\\ (3.336\%)\end{tabular}  & \begin{tabular}[c]{@{}c@{}}2.0-2.5\\ (9.769\%)\end{tabular}  & \begin{tabular}[c]{@{}c@{}}2.5-3.0\\ (18.614\%)\end{tabular} & \begin{tabular}[c]{@{}c@{}}3.0-3.5\\ (34.875\%)\end{tabular} & \begin{tabular}[c]{@{}c@{}}3.5-4.0\\ (31.857\%)\end{tabular} & \begin{tabular}[c]{@{}c@{}}4.0-4.5\\ (1.003\%)\end{tabular} & \begin{tabular}[c]{@{}c@{}}Weighted\\ Avg.\end{tabular} \\ \hline
		Gaussian & 0.0911 & \textbf{0.0387} & 0.0336 & \textbf{0.0158} & \textbf{0.0160} & 0.0360 & \textbf{0.0446} & \textbf{0.0254} \\ \hline
		Gamma & \textbf{0.0296} & 0.0456 & \textbf{0.0217} & 0.0362 & 0.0310 & 0.0852 & 0.0907 & 0.0494 \\ \hline
		Weibull & 0.0664 & 0.0597 & 0.0633 & 0.0323 & 0.0433 & \textbf{0.0249} & 0.0473 & 0.0381 \\ \hline
	\end{tabular}
	\label{tab:GoF}
\end{table*}

As shown in Fig.~\ref{fig:class_vs_iqa}, the label of the classification task and the IQA task are quite different. For the classification task, the label is one-hot encoding with the probability of certain class being 100\%. However, for the IQA task, one distorted image will receive divergent opinions from different subjects and 
people rarely reach an absolute consensus, thus resulting in the uncertainty of the label. In order to further explore the characteristics of the rating distributions of IQA, we collect images with each subject's annotation from KonIQ-10K~\cite{KonIQ-10k}. KonIQ-10K adopts the single stimulus absolute category rating method to collect the discrete rating ranging from 1 to 5. The mean opinion score (MOS) is obtained by averaging all subjects' ratings. As shown in Fig.~\ref{fig:clusters}, we first split the MOS value into 7 segments and cluster the rating distributions  of images whose mean quality score lines in a specific range. Following~\cite{murray2012ava}, we use k-means to generate 5 clusters for each quality range, and visualize the rating distributions of these clusters. From Fig.~\ref{fig:clusters}, we can see that the clusters of rating distributions tend to be approximated by Gaussian. Besides, we employ three distributions: Gaussian, Gamma and Weibull to conduct the Goodness-of-Fit (GoF) experiments respectively, and the RMSE measuring results are shown in Tabel~\ref{tab:GoF}, which reveals that Gaussian functions can perform adequately for simulating the rating distributions of IQA.


From the above preliminary experiments, we conclude that the rating distributions of IQA should follow three typical characteristics:
\begin{itemize}
    \item The uncertainty should be relatively small. All the ratings should be fluctuated around the mean quality scores. For example, a rating distribution is not expected to be like a uniform distribution with equal probabilities of each rating category.
    \item The rating distributions of all images should be diverse. Due  to the distortion diversity and the content variation, the rating distributions of different images should be diverse.
    \item The rating distributions can be closely approximated by Gaussian functions. The highest point of the distribution corresponds to the mean quality score.
\end{itemize}

Based on the above observations, we introduce three self-supervised training objectives respectively to make the predicted distributions more consistent with the human scoring nature:
\begin{itemize}
    \item Entropy loss. We encourage more confident predictions by minimizing the entropy of the predicted rating distribution for each single image.
    \item Diversity loss. We encourage the diversity of the predictions between different images by maximizing the entropy of the averaged batch predictions. The diversity loss can effectively avoid the trivial solution of entropy loss where all the unlabeled data have the same predictions that tend to be one-hot encoding.
    \item Gaussian regularization loss. We constrain that the predicted distributions should follow the Gaussian distribution.  
\end{itemize}

Apart from the self-supervised training objectives specially tailored for IQA, we also propose a more efficient and simple way to fine-tune the trained source network. Instead of fine-tuning the whole network or the whole feature extractor like most existing SFUDA methods~\cite{SHOT,ishii2021source}, we propose to re-calculate the BN statistics and only optimize the affine parameters of batch normalization (BN) layers towards target domain. Inspired by  some prior works aligning the BN statistics between the source domain and the target domain~\cite{adaptiveBN,autoBN}, we think that the affine parameters of BN can well calibrate the ``source-style'' features to ``target-style'' features. Specially, in order to avoid the conflicts between multiple target domains, we employ domain specific batch normalization (DSBN)~\cite{DSBN} to separate domain-specific information, thus making it possible to process multiple target domains simultaneously.  Besides, thanks to the DSBN, the proposed method can also be extended to an unsupervised continual learning scenario where each target domain comes sequentially.

In summary, our contributions can be summarized as follows:
\begin{itemize}
    \item We take the first step towards the source-free unsupervised domain adaptation in the IQA area, which only utilizes a trained source model and the unlabeled target domain data to solve UDA problems.
    
    \item We design three self-supervised training objectives according to three observed IQA intrinsic properties, and demonstrate that simply fine-tuning the affine parameters of the BN layers can achieve efficient and fast adaptation to the target domain.
    
    \item We use domain-specific batch normalization layer to avoid the conflicts between different domains, which makes it possible to process multiple target domains simultaneously. Besides, the proposed method can also be extended to unsupervised continual learning scenario.
\end{itemize} 

\section{Related works}
\subsection{Blind image quality assessment}

BIQA aims to automatically predict the subjective quality of a distorted image without accessing any reference information. It can be roughly divided into two categories: distortion-specific and general-purpose approaches ~\cite{zhai2020perceptual}. Distortion-specific methods are designed for a particular distortion type (\textit{e.g.} JPEG ~\cite{JPEG-BIQA}, blur~\cite{BLUR-BIQA} and super-resolution~\cite{SR-BIQA}). General-purpose methods design models based on the assumption that distortions will break the regular statistics extracted from natural images~\cite{NIQE,liu2014no}.
Recently, deep learning has significantly advanced the development of BIQA.
Typically, learning-based BIQA methods~\cite{zhou2019dual,xu2020blind} usually employ the two-step network, \textit{i.e.},  feature extraction and quality prediction. 
Pan \textit{et al}.~\cite{pan2018blind} employed a fully convolutional neural network to predict a pixel-by-pixel similar quality map and used a deep pooling network to regress the quality map into a quality score. In~\cite{zhang2018blind}, Zhang \textit{et al.} proposed a deep bilinear model that works for both synthetically and authentically distorted images. 

It is known that learning-based BIQA models are
vulnerable to cross-distortion scenario generalization due to the limited number of images with subjective ratings, and therefore some works are proposed to alleviate the poor generalizability of BIQA models. RankIQA~\cite{rankiqa} ranked images in terms of image quality by using synthetically generated distortions, to address the problem of limited IQA dataset size. Zhang \textit{et al.}~\cite{zhang2021uncertainty} proposed an uncertainty-aware BIQA model and a method of training it on multiple datasets simultaneously. Liu \textit{et al.}~\cite{liu2021liqa} and Zhang \textit{et al.}~\cite{delange2021continual} attempted to solve the continual learning of BIQA. UCDA~\cite{uncertaintyDA} and RankDA~\cite{rankDA} employed the unsupervised domain adaptation technique to transfer knowledge from the label-sufficient source domain to the label-free target domain. Although the above UDA-based methods have achieved impressive performance, they are all designed to utilize the source data when adapting the model, which is impractical when the source data is unavailable due to storage or private issues. In contrast, our work tries to investigate how to achieve a more practical and flexible adaptation without access to the source data. 
\begin{figure*}[htbp]
\centering
\includegraphics[width=0.7\textwidth]{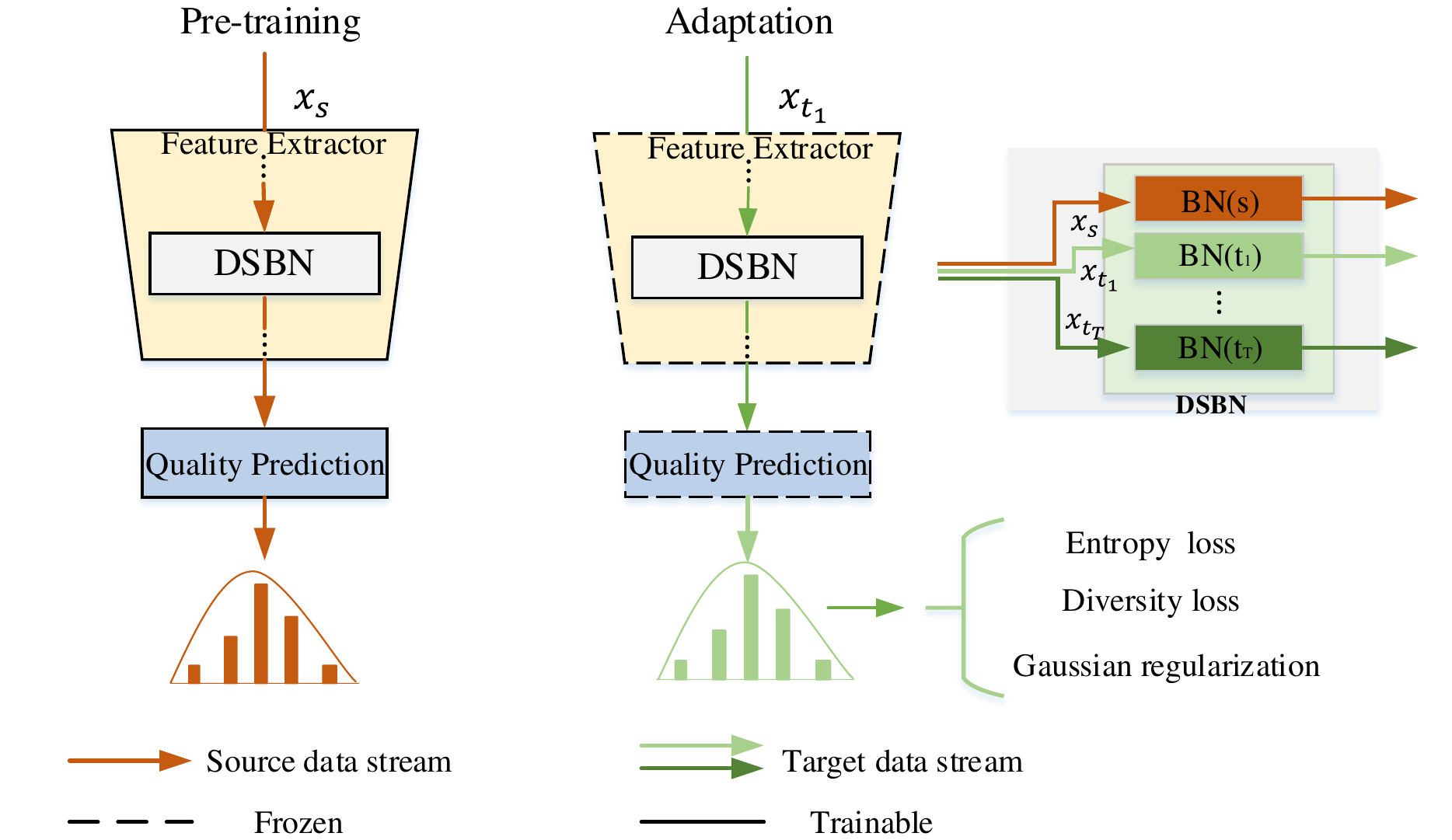}
\caption{Framework of our proposed method. The dashed line means that the parameters are frozen while the solid line means that the parameters are trainable. During the adaptation stage, we simply optimize the affine parameters of the BN layers for each target domain.}
\label{fig:network arch}
\end{figure*}
\subsection{Unsupervised domain adaptation}
Unsupervised domain adaptation (UDA) is proposed to tackle the domain shift when the labeled source data and the unlabeled target data fall from different distributions. The goal of UDA is to align the distributions between the source and the target domains. The mainstream UDA methods can be categorized into two directions: discrepancy-based approaches and adversary-based approaches. Discrepancy-based approaches \cite{long2015learning,zhang2015deep} 
mostly use distance measures such as maximum mean dispcrepancy (MMD) \cite{gretton2006kernel}, Wasserstein metric, correlation alignment (CORAL)~\cite{sun2017correlation}, Kullback-Leibler (KL) divergence \cite{kullback1951information}, and contrastive domain discrepancy (CDD) \cite{kang2019contrastive} to measure the dissimilarity across domains. 
Adversary-based approaches\cite{ganin2015unsupervised,pei2018multi} tend to obtain transferable and domain-invariant features through an adversarial objective.
However, the source data is not always available due to storage or privacy isssues in many
real-world applications, which makes the above UDA approaches unable to use.
To address this issue, source-free UDA is proposed, which adapts to the target domain only with target data and the model pre-trained on source domain provided. SHOT\cite{SHOT} proposed an information maximization loss and a clustering-based pseudo labeling method to implicitly align representations from the target domains to the source hypothesis. Ishii \textit{et al.}~\cite{ishii2021source} proposed to minimize both the BN-statistics matching loss and the information maximization loss to fine-tune the feature extractor. TENT\cite{wang2020tent} proposed to optimize the BN affine parameters by simply minimizing the entropy of the predictions during test-time adaptation.
Although several works have attempted to utilize UDA technique to solve the domain shift in IQA , the source-free UDA in IQA area is still undex-explored, which is an interesting but challenging topic. 

\section{Problem definition}
For a vanilla UDA task, we are usually given $n_{s}$ labeled samples $\{\mathbf{x}_{s}^{i},y_{s}^{i}\}_{i=1}^{{{n}_{s}}}$ from the source domain $\mathcal{D}_{s}$ where $\mathbf{x}_{s}^{i}\in {{\mathcal{X}}_{s}}$, $y_{s}^{i}\in {{\mathcal{Y}}_{s}}$, and $n_{t}$ unlabeled samples $\{\mathbf{x}_{t}^{i}\}_{i=1}^{{{n}_{t}}}$ from the target domain $\mathcal{D}_{t}$, where $\mathbf{x}_{t}^{i}\in {{\mathcal{X}}_{t}}$. The goal of vanilla UDA is to predict the target labels $\{y_{t}^{i}\}_{i=1}^{{{n}_{t}}}$, where $y_{t}^{i}\in {{\mathcal{Y}}_{t}}$. We denote the source task and target task as ${{\mathcal{X}}_{s}} \to {{\mathcal{Y}}_{s}}$ and ${{\mathcal{X}}_{t}} \to {{\mathcal{Y}}_{t}}$, respectively. While for the source-free UDA, we are expected to learn a target function $f_{t}:{{\mathcal{X}}_{t}} \to {{\mathcal{Y}}_{t}}$ and infer $\{y_{t}^{i}\}_{i=1}^{{{n}_{t}}}$ with only a pre-trained source function $f_{s}:{{\mathcal{X}}_{s}} \to {{\mathcal{Y}}_{s}}$ and the unlabeled target data $\{\mathbf{x}_{t}^{i}\}_{i=1}^{{{n}_{t}}}$ provided.

\section{Approach}
In this section, we first illustrate how to cast the quality assessment task to a rating distribution prediction task. Then we introduce the source-free adaptation process of our proposed method.

\subsection{Discrete rating distribution construction and learning}
To bridge the gap between the classfication and the IQA, we aim to cast the quality asessment task to a discrete rating distribution prediction task, which is helpful to transfer the technologies from classification to IQA. Given the mean quality score $\mu$ and the variance $\sigma^2$ of a single image, we first model the rating distribution as a truncated Gaussian distribution \cite{burkardt2014truncated} as Eq. ~\ref{equ:Truncated gaussian}:
\begin{equation}
    p(l|\mu ,{{\sigma }^{2}})=\frac{\varphi (\mu ,{{\sigma }^{2}};l)}{\int_{{{r}_{1}}}^{{{r}_{2}}}{\varphi (\mu ,{{\sigma }^{2}};l)dl}},r_{1}<=l<=r_{2}
    \label{equ:Truncated gaussian} 
\end{equation}
where $l$ is the variable of the quality score, $r_{1}$ and $r_{2}$ are the lower and upper bounds of the quality range. $\varphi$ is the standard normal distribution where $\varphi (\mu ,{{\sigma }^{2}};l)=\frac{1}{\sqrt{2\pi }\sigma }\exp \left( -\frac{{{(l-\mu )}^{2}}}{2{{\sigma }^{2}}} \right)$. To make this distribution discrete, we sample $C$ quality levels at equal intervals, where $C$-level is crossponding to $C$ classes in classification. The $k$-th rating level $l_k$ is:
\begin{equation}
    l_k = \frac{k-1}{C-1}*(r_{2}-r_{1})+r_{1}, k\in[1,C]
    \label{equ:discrete}
\end{equation}
Then we construct the discrte rating distribution $\mathbf{q}=(q_{1},\cdots,q_{C})$ using:
\begin{equation}
{{q}_{k}}=\frac{p({{l}_{k}}|\mu,\sigma^2 )}{\sum\nolimits_{j}{p({{l}_{j}}|\mu, \sigma^2 )}},
\end{equation}
where $\sum\limits_{k=1}^{C}{{{q}_{k}}}=1$. $L=(l_{1},\cdots,l_{C})$ is the rating level set and $\sum\limits_{k=1}^{C}{{{q}_{k}}}{{l}_{k}}\approx \mu$.
\begin{table*}[htbp]
\centering
\caption{Description of IQA databases. DisNum refers to the number of synthetic distortion types. MOS refers to Mean Opinion Score and a higher value denotes better perceptual quality. DMOS refers to Differential Mean Opinion Score and is inversely proportional to MOS. DisImageNum refers to the number of distorted images.}
\scalebox{0.92}{
\begin{tabular}{c|cccccc}
\hline
Database & Scenario & DisNum & Subjective Testing Methodology & Annotation & Range & DistImageNum  \\ \hline
LIVE\cite{LIVE} & Synthetic & 5 & Single stimulus continuous quality rating & DMOS,Variance & {[}0,100{]} & 779  \\
CSIQ\cite{CSIQ} & Synthetic & 6 & Multi stimulus absolute category rating & DMOS,Variance & {[}0,1{]} & 866  \\
KADID-10K\cite{KADID-10K} & Synthetic & 25 & Double stimulus absolute category rating with crowdsourcing & MOS,Variance & {[}1,5{]} & 10,125  \\ \hline
BID\cite{BID} & Authentic & - & Single stimulus continuous quality rating & MOS,Variance & {[}0,5{]} & 586 \\
KonIQ-10K\cite{KonIQ-10k} & Authentic & - & Single stimulus absolute category rating with crowdsourcing & MOS,Variance & {[}1,5{]} & 10,073 \\ \hline
\end{tabular}}
\label{tab:dataset}
\end{table*}
A deep IQA network is usually composed of two parts: feature extractor and quality prediction head (several fully-connected layers). Given an input image $\mathbf{x}$ with ground truth mean quality score $\mu$ and  rating distribution $\mathbf{q}$, the activation of the last fully connected layer is $\mathbf{z}=\phi(\mathbf{x};\mathbf{\theta})=(z_{1},\cdots,z_{C})$, where $\mathbf{\theta}$ is the parameters of the IQA network $\phi$. We use a softmax function to transform $\mathbf{z}$ into a probabiliy distribution:
\begin{equation}
    {{\hat{q}}_{k}}=\frac{\exp ({{z}_{k}})}{\sum\nolimits_{j}{\exp ({{z}_{j}})}}.
\end{equation}
The goal is to find $\mathbf{\theta}$ to generate a distribution $\hat{\mathbf{q}}$ that is similar to $\mathbf{q}$ with the predicted mean quality score $\hat{\mu}$ keeps close to $\mu$.
During training on the source domain, we construct the following learning objective:
\begin{equation}
    \begin{aligned}
  & \mathcal{L}={{\mathcal{L}}_{KL}}+{{\mathcal{L}}_{MSE}}= \\ 
 & {{\mathbb{E}}_{\{\mathbf{x}^{i}\}_{i=1}^{B}\in {{\mathcal{X}}_{s}}}} \left[-\sum\limits_{k}{{{q}_{k}}\log }{{{\hat{q}}}_{k}}+{{\left\| \mu-\hat{\mu} \right\|}_{2}}\right],  
\end{aligned}
\label{equ:source loss}
\end{equation}
where $\hat{\mu}=\sum\limits_{k=1}^{C}{{{{\hat{q}}}_{k}}}{{l}_{k}}$ and B is the mini-batch size. ${{\mathcal{L}}_{KL}}$ minizes Kullback-Leibler (KL) divergence between the predicted distribution and the ground-truth distribution. In this way, we cast the quality assessment task as a rating distribution prediction task.

\begin{figure*}[t!]
\centering
\includegraphics[scale=0.5]{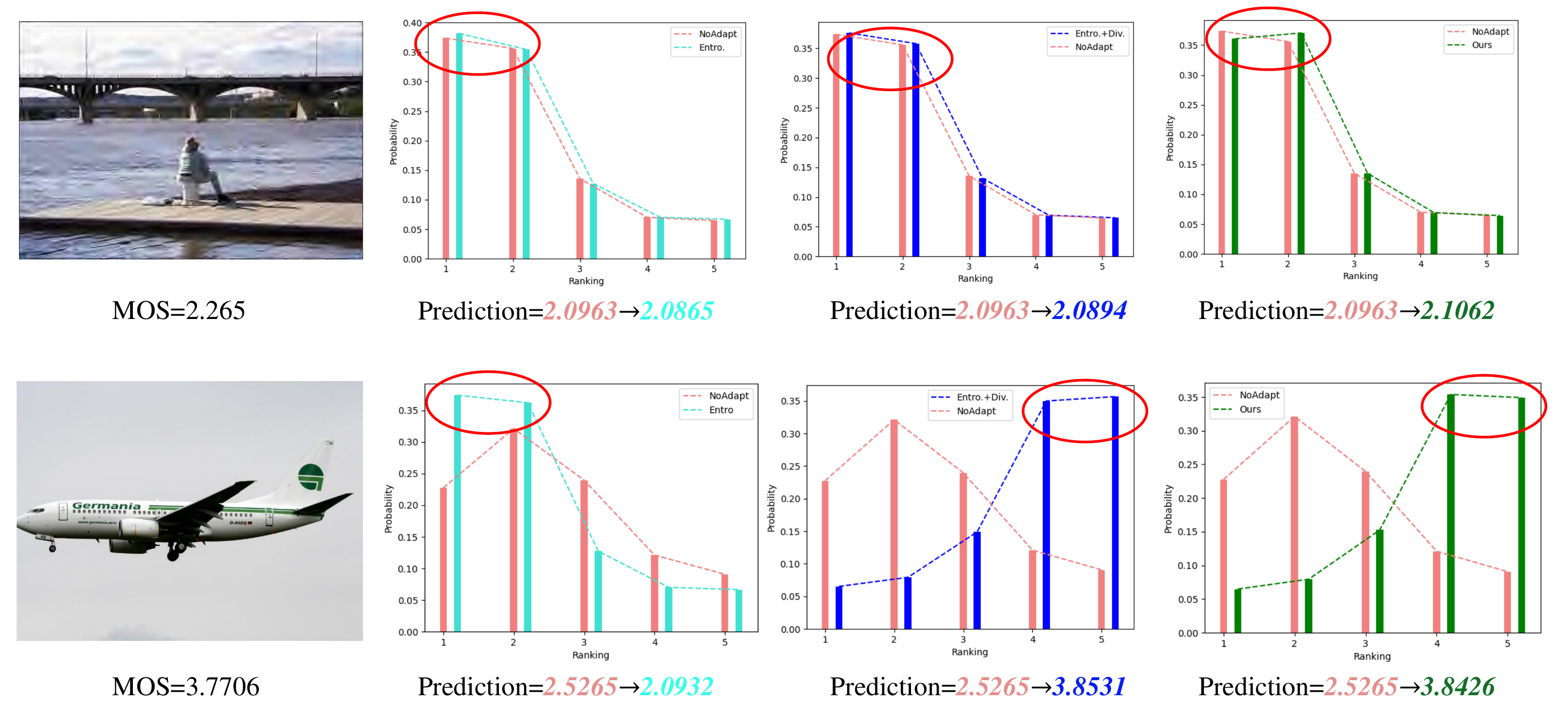}
\caption{Comparison between differnent combinations of objectives. Adaptation results using \textit{Entro.}, \textit{Entro.+Div.} and \textit{Entro.+ Div.+Gau.} are shown from left to right, respectively. Our method can produce more accurate results consistent with the nature of human scoring (Guassian distribution). To avoid overlap, we shift the x-coordinate a little for adapted results.}
\label{fig:sample objective}
\end{figure*}

\subsection{Adaptation to target domain}

As shown in Fig.~\ref{fig:network arch}, during the adaptation process, we freeze the parameters of all the network except for the affine parameters within the domain-specific batch normalization (DSBN) layers. DSBN is just like the ordinary BN, but can work for multiple domains simultaneously.
A BN layer whitens activations within a mini-batch of $B$ examples for each channel dimension and transforms the whitened activations using affine parameters $\gamma$ and $\beta$. DSBN allocates domain-specific affine parameters $\gamma_{d}$ and $\beta_{d}$ for each domain label $d\in\{s,t_{1},\cdots,t_{T}\}$. Let $\mathbf{v}_{d}\in\mathbb{R}^{B\times H\times W}$ denote the activations of each channel from domain $d$, $B$ is the mini-batch size and $H\times W$ represents the spatial size. DSBN can be defined as:
\begin{equation}
    DSB{{N}_{d}}({{\mathbf{v}}_{d}}[b,i,j];{{\gamma }_{d}},{{\beta }_{d}})={{\gamma }_{d}}\cdot {{\hat{\mathbf{v}}}_{d}}[b, i,j]+{{\beta }_{d}},
\end{equation}
where  ${{\hat{\mathbf{v}}}_{d}}[b,i,j]$ is the whitened activations:
\begin{equation}
    {{\hat{\mathbf{v}}}_{d}}[b,i,j]=\frac{{{\mathbf{v}}_{d}}[b,i,j]-{{{\tilde{\mu }}}_{d}}}{\sqrt{\tilde{\sigma }_{d}^{2}+\varepsilon }}.
\end{equation}
$\varepsilon$ is a small constant to avoid divide-by-zero, and ${{\tilde{\mu }}_{d}}$ and ${{\tilde{\sigma }}_{d}}$ are statistics (the mean and variance) of activations within a mini-batch:
\begin{equation}
    {{\tilde{\mu }}_{d}}=\frac{\sum\nolimits_{b}{\sum\nolimits_{i,j}{{{\mathbf{v}}_{d}}[b,i,j]}}}{B\cdot H\cdot W},
\end{equation}

\begin{equation}
    \tilde{\sigma }_{d}^{2}=\frac{\sum\nolimits_{b}{\sum\nolimits_{i,j}{({{\mathbf{v}}_{d}}[b,i,j]}}-{{{\tilde{\mu }}}_{d}}{{)}^{2}}}{B\cdot H\cdot W}.
\end{equation}
During training, DSBN estimates the mean and variance of activations for each domain separately by exponential moving average with update factor $\alpha$:
\begin{equation}
    {{\bar{\tilde{\mu }}}_{d}}^{ite+1}=(1-\alpha ){{\bar{\tilde{\mu }}}_{d}}^{ite}+\alpha {{\tilde{\mu }}_{d}}^{ite+1}.
\end{equation}
\begin{equation}
    {{(\bar{\tilde{\sigma }}_{d}^{ite+1})}^{2}}=(1-\alpha ){{(\bar{\tilde{\sigma }}_{d}^{ite})}^{2}}+\alpha {{(\tilde{\sigma }_{d}^{ite+1})}^{2}},
\end{equation}
where during testing stage, ${{\bar{\tilde{\mu }}}_{d}}$ and ${{(\bar{\tilde{\sigma }}_{d})}^{2}}$ are directly used for the testing examples in each domain $d$.

For the $\tau$-th target domain $t_{\tau}$, we update the BN statistics and optimize the affine-parameters $\{\gamma_{t_{\tau},m,h},\beta_{t_{\tau},m,h}\}$ for each BN layer $m$ and channel $h$ in the DSBN, where $\tau\in\{1,\cdots,T\}$. According to the three observed characteristics of IQA distributions, we design three self-supervised objectives: Enropy loss, Diversity loss and Gaussian regularization loss.

\textbf{Entropy loss.} Based on the observation that all the ratings are fluctuated around the mean quality scores, we minimize the entropy of the predicted rating distribution for each single image to encourage more confident predictions:
\begin{equation}
  {{\mathcal{L}}_{ent}}=-{{\mathbb{E}}_{\{\mathbf{x}^{i}\}_{i=1}^{B}\in {{\mathcal{X}}_{t_{\tau}}}}}\sum\limits_{k=1}^{C}{\hat{q}_{k}^{i}}\log \hat{q}_{k}^{i}
\end{equation}
where ${{\hat{q}}^{i}}=\delta (\phi (\mathbf{x}^{i};\theta ))$ is the softmax output of the IQA network $\phi$ with parameters $\theta$. $\delta$ is the softmax funtion and $B$ is the mini-batch size.

\textbf{Diversity loss.} Based on the observation that the rating distritions of different images are diverse, we maximize the entropy of the averaged prediction within a mini-batch to make the results globally diverse:
\begin{equation}
   {{\mathcal{L}}_{div}}=-\sum\limits_{k=1}^{C}{{{{\bar{q}}}_{k}}}\log {{\bar{q}}_{k}},
\end{equation}
where $\bar{q}={{E}_{\{\mathbf{x}^{i}\}_{i=1}^{B}\in {{\mathcal{X}}_{t_{\tau}}}}}[\delta (\phi (\mathbf{x}^{i};\theta ))]$ is the averaged prediction within a mini-batch.

\textbf{Gaussian regularization loss.} Based on the observation that the rating distributions can be closedly approximately by Gaussian functions, we design the Gaussian regularization loss to encourage the predicted predictions to be more consistent with the nature of human scoring:
\begin{equation}
    {{\mathcal{L}}_{Gau}}=-{{E}_{\{\mathbf{x}^{i}\}_{i=1}^{B}\in {{\mathcal{X}}_{t_{\tau}}}}}\sum\limits_{k=1}^{C}{\tilde{q}_{k}^{i}\log \hat{q}_{k}^{i}},
\end{equation}
where $\tilde{\mathbf{q}}$ is the pseudo rating distribution:
\begin{equation}
    {{\tilde{q}}_{k}}=\frac{p({{l}_{k}}|\hat{\mu},\hat{\sigma }^2)}{\sum\nolimits_{j}{p({{l}_{j}}|\hat{\mu},\hat{\sigma }^2)}}.
\end{equation}
$p$ is the truncated normal distribution defined in Eq.~\ref{equ:Truncated gaussian}.
$\hat{\mu}$ and $\hat{\sigma}^2$ are the estimated mean and variance of the prediced rating distribution respectively. $\hat{\mu}=\sum\limits_{k=1}^{C}{{{{\hat{q}}}_{k}}{{l}_{k}}}$ and ${{\hat{\sigma }}^{2}}=\sum\limits_{k=1}^{C}{{{{\hat{q}}}_{k}}}{{({{l}_{k}}-\hat{\mu})}^{2}}$

The total loss on $T$ target domains can be defined by:
\begin{equation}
    {{\mathcal{L}}_{total}}=\frac{1}{T}\sum\limits_{\tau }{(\mathcal{L}_{ent}^{\tau }-\lambda _{div}^{\tau }\mathcal{L}_{div}^{\tau }+\lambda _{Gau}^{\tau }\mathcal{L}_{Gau}^{\tau })},
    \label{equ:target loss}
\end{equation}
where $\lambda _{div}^{\tau }$ and $\lambda _{Gau}^{\tau }$ are hyperparameters of $\tau$-th target domain which balance the weights of corresponding loss items.

\section{Experiments}
In this section, we first introduce the IQA datasets and the implementation details of our experiments. Then, we compare our method with several state-of-the-art methods and conduct ablation studies to verify the effectiveness of each item of our designed self-supervised loss. Finally, we show the performance of our method under the unsupervised continual learning setting.
\subsection{Datasets}
\label{sec:datasets}
We conduct cross-domain experiments on five IQA datasets, among which three are synthetically distorted (LIVE \cite{LIVE}, CSIQ \cite{CSIQ}, KADID-10K \cite{KADID-10K}) and the others are authentically distorted (BID \cite{BID} and KonIQ-10K \cite{KADID-10K}). The summarization of the five datasets is shown in Table \ref{tab:dataset}.

\textbf{LIVE} database includes 779 synthetically distorted images, which are generated from 29 reference images by corrupting them with five distortion types. DMOS of each distorted image ranges from 0 to 100 and is collected using the single stimulus continuous quality rating method. \textbf{CSIQ} database consists of 866 synthetically distorted images which are derived from 30 original images distorted with six distortion types at four to five different intensity levels. The ratings are reported in the form of DMOS ranging from 0 to 1 using multi-stimulus absolute category rating method. \textbf{KADID-10K} consists of 10,125 distorted images derived from 81 pristine images considering 25 different distortion types at 5 intensity levels. The MOS of each image ranges from 1 to 5 and is collected using double stimulus absolute category rating with crowdsourcing. \textbf{BID} database contains 586 authentically distorted pictures taken by human users in a variety of scenes, camera apertures, and exposition times. The distorted images are mostly blurred, which include not only  typical, easy-to-model blurring cases but also more complex, realistic ones. The MOS of each image ranges from 0 to 5 and is collected using the single stimulus continuous rating method.
\textbf{KonIQ-10K} database consists of 10,073 authentically distorted images selected from a massive public multimedia database, YFCC100m \cite{YFCC100M}.  The MOS of each image ranges from 1 to 5 and is collected by the single stimulus absolute category rating method with crowdsourcing.
\begin{table*}[]
\centering
\caption{SROCC and PLCC performance comparisons of different algorithms under various cross-domain scenarios.}
\begin{tabular}{c|ccccccc}
\hline
Source & \multicolumn{2}{c}{{\diagbox{Alg.}{SROCC/PLCC}{Target}}} & KADID-10K & KonIQ-10K & BID & CSIQ & LIVE \\ \hline
\multirow{13}{*}{KADID-10K} & \multicolumn{1}{c|}{\multirow{3}{*}{Trad.}} & \multicolumn{1}{c|}{NIQE\cite{NIQE}} & 0.3464/0.3792 & 0.4469/0.4600 & 0.3640/0.5046 & 0.4895/0.6645 & 0.8838/0.8876 \\
 & \multicolumn{1}{c|}{} & \multicolumn{1}{c|}{PIQE\cite{PIQE}} & 0.2424/0.3325 & 0.0843/0.1995 & 0.2376/0.2437 & 0.5512/0.6721 & 0.7787/0.7881 \\
 & \multicolumn{1}{c|}{} & \multicolumn{1}{c|}{BRISQUE\cite{BRISQUE}} & 0.2890/0.3704 & 0.3910/0.3815 & 0.2494/0.3341 & 0.5853/0.5912 & \textbf{0.9456/0.9399} \\ \cline{2-3}
 & \multicolumn{1}{c|}{\multirow{4}{*}{Lean.}} & \multicolumn{1}{c|}{DBCNN\cite{DBCNN}} & 0.8603/0.8666 & 0.5032/0.4584 & 0.5231/.5252 & 0.7408/0.6673 & 0.8658/0.8296 \\
 & \multicolumn{1}{c|}{} & \multicolumn{1}{c|}{HyperIQA\cite{HyperIQA}} & 0.8416/0.8429 & 0.4259/0.3855 & 0.3792/0.2937 & 0.7334/0.6699 & 0.8936/0.8304 \\
  & \multicolumn{1}{c|}{} & \multicolumn{1}{c|}{NIMA\cite{NIMA}} & 0.8400/0.8521 & 0.5876/6017 & 0.4817/0.4933 & 0.7455/0.7748 & 0.8262/0.8489 \\
 & \multicolumn{1}{c|}{} & \multicolumn{1}{c|}{RankIQA\cite{rankiqa}} & 0.7406/0.7335 & 0.3517/0.2884 & 0.2292/0.1521 & 0.6298/0.6109 & 0.8063/0.8082 \\
 & \multicolumn{1}{c|}{} & \multicolumn{1}{c|}{MetaIQA\cite{metaiqa}} & 0.8156/0.8246 & 0.5272/0.6362 & 0.2121/0.2017 & 0.5515/0.7077 & 0.8667/0.9071 \\ \cline{2-3}
 & \multicolumn{2}{c|}{NoAdapt} & \multirow{6}{*}{0.8355/0.8372} & 0.6371/0.6233 & 0.5378/0.5084 & 0.7452/0.7666 & 0.8783/0.8847 \\ \cline{2-3}
 & \multicolumn{1}{c|}{\multirow{2}{*}{UDA}} & \multicolumn{1}{c|}{UCDA\cite{uncertaintyDA}} &  & 0.4958/0.5010 & 0.5499/0.5277 & 0.6794/0.7189 & 0.8969/0.8919 \\
 & \multicolumn{1}{c|}{} & \multicolumn{1}{c|}{RankDA\cite{rankDA}} &  & 0.6364/0.5662 & 0.5273/0.578 & 0.7561/0.6402 & 0.7152/0.7024 \\ \cline{2-3}
 & \multicolumn{1}{c|}{\multirow{3}{*}{SFUDA}} & \multicolumn{1}{c|}{TENT\cite{wang2020tent}} &  & 0.6735/0.6004 & \textbf{0.6186/0.6200} & 0.7482/0.7752 & 0.8515/0.8556 \\
 & \multicolumn{1}{c|}{} & \multicolumn{1}{c|}{SHOT\cite{SHOT}} &  & 0.6839/0.6937 & 0.5622/0.5798 & 0.7588/0.7312 & 0.8553/0.8634 \\
 & \multicolumn{1}{c|}{} & \multicolumn{1}{c|}{Ours} &  & \textbf{0.7221/0.7124} & 0.5946/0.5982 & \textbf{0.7658/0.7855} & 0.8942/0.8867 \\ \hline
\multirow{2}{*}{KonIQ-10K} & \multicolumn{2}{c|}{NoAdapt} & 0.5518/0.5821 & \multirow{2}{*}{0.8654/0.8784} & 0.7511/0.7645 & 0.7105/0.7608 & 0.7266/0.7226 \\
 & \multicolumn{2}{c|}{Ours} & 0.5829/0.6024 &  & 0.7574/0.7766 & 0.7699/0.8121 & 0.8577/0.8444 \\ \hline
\multirow{2}{*}{BID} & \multicolumn{2}{c|}{NoAdapt} & 0.3761/0.4371 & 0.5751/0.6192 & \multirow{2}{*}{0.8082/0.7805} & 0.4481/0.4355 & 0.6159/0.5762 \\
 & \multicolumn{2}{c|}{Ours} & 0.3960/0.4606 & 0.6552/0.6579 &  & 0.4743/0.4432 & 0.6259/0.5933 \\ \hline
\multirow{2}{*}{CSIQ} & \multicolumn{2}{c|}{NoAdapt} & 0.5214/0.5836 & 0.6182/0.6118 & 0.5598/0.5447 & \multirow{2}{*}{0.9608/0.9472} & 0.9107/0.9024 \\
 & \multicolumn{2}{c|}{Ours} & 0.5243/0.5973 & 0.6669/0.6744 & 0.5935/0.5996 &  & 0.9053/0.9035 \\ \hline
\multirow{2}{*}{LIVE} & \multicolumn{2}{c|}{NoAdapt} & 0.4768/0.5230 & 0.7210/0.7387 & 0.6718/0.6926 & 0.7223/0.8003 & \multirow{2}{*}{0.9731/0.9573} \\
 & \multicolumn{2}{c|}{Ours} & 0.5319/0.5387 & 0.7165/0.7146 & 0.6983/0.6799 & 0.7502/0.8025 & \\ \hline
\end{tabular}
\label{tab:comparison}
\end{table*}


\begin{table*}[t]
\centering
\caption{SROCC/PLCC performance with respect to differnent combinations of the entropy loss, diversity loss and the Gaussian regularization loss.↑ and ↓ mean performance improvement and degradation compared with NoAdapt results.}
\begin{tabular}{c|cccc}
\hline
SROCC/PLCC & KonIQ-10K & BID & CSIQ & LIVE \\ \hline
NoAdapt & 0.6378/0.6252 & 0.5400/0.5101 & 0.7452/0.7666 & 0.8783/0.8847 \\
Entro. & 0.6812/0.6731↑ & 0.6186/0.6200↑ & 0.7482/0.7752↑ & 0.8515/0.8556↓ \\
Div. & 0.6561/0.6593↑ & 0.5060/0.4915↓ & 0.7480/0.7770↑ & 0.8936/0.8816↑ \\
Gau. & 0.6212/0.6161↓ & 0.5351/0.5449↓ & 0.7449/0.7625↓ & 0.9112/0.9027↑ \\
Entro.+Div. & 0.7058/0.6989↑ & 0.5989/0.6596↑ & 0.7550/0.7704↑ & 0.8431/0.8509↓ \\
Entro.+Gau. & 0.6748/0.6624↑ & 0.6212/0.6215↑ & 0.7614/0.7828↑ & 0.8599/0.8580↓ \\
Div.+ Gau. & 0.6005/0.5897↓ & 0.5281/0.5346↓ & 0.7542/0.7820↑ & 0.9090/0.9082↑ \\ \hline
Ours & 0.7221/0.7124↑ & 0.5946/0.5982↑ & 0.7658/0.7855↑ & 0.8942/0.8867↑ \\\hline
\end{tabular}
\label{tab:ablation}
\end{table*}

\subsection{Implementation details}
To evaluate the performance of our proposed method under cross-domain scenario, we utilize the above 5 datasets to construct $5\times4$ cross-domain settings. For each dataset, we randomly sampled 80\% images for training, 10\% for validation and the left 10\% for testing. Specially, for the synthetic datasets, we split the datasets according to the reference images in order to ensure content dependence.  During training we randomly cropped the images into 384 × 384, and during validation and testing, we cropped the images to 384 × 384 in the center. We set the quality range lower bound $r_{1}$ and the upper bound $r_{2}$ to 1 and 5, respectively. Then we linearly rescaled the MOS/DMOS of each image to the range of [1-5] for all datasets, and the variance was also correspondingly scaled. We empirically set the number of sampled rating levels $C$ to 5.

We employ a pre-trained ResNet-18 (without the final $FC$ layers) as the feature extractor and use two $FC-ReLU$ layers followed by a softmax function as the quality prediction head. We emprically set $\lambda_{div}$ to 1.0 and $\lambda_{Gau}$ to 0.2 for all target domains in all experiments.  When training on the source domain, the Adam optimizer\cite{adam} with a leaning rate of $1e^{-4}$ was adopted to minimize the loss defined in Eq.~\ref{equ:source loss}.  We adopted an early-stopping strategy and chose the model that performed the best on the validation set as the pre-trained source model. 
During the adaptation stage, we employed the Adam optimizer with a leaning rate of $5e^{-5}$  to minimize the loss defined in Eq.~\ref{equ:target loss}. We choose the adapted model which achieves the lowest loss for testing. We adopted two performance criteria: Spearman rank-order correlation coefficient (SROCC) and Pearson linear correlation
coefficient (PLCC), to measure prediction monotonicity and precision, respectively. Before computing PLCC, the predicted quality scores are passed through a non-linear logistic mapping function:
\begin{equation}
    \hat{\mu}={{\beta }_{1}}(\frac{1}{2}-\frac{1}{\exp ({{\beta }_{2}}(\hat{\mu}-{{\beta }_{3}}))})+{{\beta }_{4}}\hat{\mu}+{{\beta }_{5}}
\end{equation}
We repeated 5 times for all experiments with different seeds and got the average values of PLCC and SROCC for report. 
\vspace{-0.4cm}
\subsection{Performance evaluation}
In order to validate the effectiveness of our proposed method under cross-domain scenario, we utilize the 5 datasets described in Section~\ref{sec:datasets} to construct $5\times4$ cross-domain settings, covering synthetic $\to$ synthetic, synthetic $\to$ authentic, authentic $\to$ synthetic and  authentic $\to$ authentic. We compare the performance of ours against three traditional methods (NIQE\cite{NIQE}, PIQE\cite{PIQE}, and BRISQUE\cite{BRISQUE}), five learning-based methods (DBCNN\cite{DBCNN}, HyperIQA\cite{HyperIQA}, RankIQA\cite{rankiqa}, MetaIQA\cite{metaiqa} and NIMA\cite{NIMA}), two UDA-based methods for quality assessment (UCDA\cite{uncertaintyDA} and RankDA\cite{rankDA}) and two SUDUA methods adapted from the classification task (TENT\cite{wang2020tent} and SHOT\cite{SHOT}). One should be noted that among the learning-based methods, NIMA also predicts the rating distributions instead of the quality score, but it approximates the distributions from the available MOS values through
maximum entropy optimization which is not consistent with human scoring nature (when the MOS value is equal to 3, the approximated distribuion is a uniform distribution where the probability of each rating category is equal to 0.2.). We evaluate the competitors' performance under the same experimental settings as ours, and the results are shown in Table~\ref{tab:comparison}. From the results, we have sevaral observations:
\begin{itemize}
    \item[(1)] All algorithms, no matter traditional methods or learning-based methods, will suffer from performance degrdation when coming across domain shift.
    \item[(2)] Comparing the ``NoAdapt'' results of ours against other methods, we can find that due to the rating distribution construction and learning, our method can achieve compraratively better generalization ability under cross-dataset setup.
    \item[(3)] Compared with UDA-based IQA methods, ours can steadily achieve better performance. In the second stage of UCDA, it tries to align the confident subdomain and uncertain domain, which can be seen as implicit entropy minimization. In RankDA, it aligns the rank feature which reveals  the pairwise quality relationship. However, it selectes the rank pairs according to the inaccurate pseudo scores given by DBCNN\cite{DBCNN} and the final quality scores are generated from inaccurate rank mos, thus it does not bring appparent improvement either.
    \item[(4)] Compared with SFUDA-based methods adapted from classification, ours can also obtain superior performance due to the objectives specially designed accoring to IQA intristic propertis. 
    \item[(5)] Our method can effctively mitigate the domain-shift under most cross-domain scenarios (\textit{e.g.} the SROCC and PLCC are improved by 0.1311 and 0.1178 respectively under KonIQ-10K $\to$ LIVE scenario).
\end{itemize}

\vspace{-2mm}
\subsection{Ablation study}
To verify the effectiveness of our designed self-supervised objectives, we conduct ablation studies under the ``KADID-10K $\to$ Others'' cross-domain scenario. We design 6 variants combinations of the entropy loss, diversity loss and the Gaussian regularization loss. The SROCC/PLCC performance are shown in Table~\ref{tab:ablation}. From the table, we can see that each single objective can improve the performance of certain datasets. Besides, the combinations of the three objectives can obtain the best performance where the performances of all datasets get improved. To intuitively understand what each objective does, we visualize the predicted distributions as well as the corresponding quality scores of some samples in Fig.~\ref{fig:sample objective}. Adaptation results using \textit{Entro.}, \textit{Entro.+Div.} and \textit{Entro.+ Div.+Gau.} are shown from left to right respectively. We can see that entropy loss is aimed to encourage more confident predictions by improving the probability of certain ranking category, but it will limit the predicted score around the initial quality score. At the same time, diversity loss will explore more diverse results and thus will probably change the peak of distribution. For example, in the second row of Fig.~\ref{fig:sample objective}, the initial predicted quality score is 2.5265.  When applying entropy loss, the distribution is sharper, but the predicted score gets worse. In contrast, when applying diversity loss, the peak of the predicted distribution shifts to the region near ranking 4. However,  by compaing the circled segments, we can find the the results produced by \textit{Entro.} and  \textit{Entro.+Div.} actually don't obey the Gaussian distribution (the mean score does not corresponds to the peak). From the fourth column of Fig.~\ref{fig:sample objective},Kon we can find that when applying the gaussian regularization loss, the predicted distribution is well calibrated and tends to approximately be a Gaussian distribution. Meanwhile, the predicted score gets more accurate compared with simply using entropy loss and diversity loss.

\vspace{-2mm}
\subsection{Extension to unsupervised continual learning setting}
Thanks to the DSBN layers, our method not only can process multiple target domains simultaneously, but also can be extended to process multiple domains sequentially which is called a continual learning setting. We sequentially learn multiple datasets following the order of 
LIVE$\to$CSIQ$\to$BID$\to$CLIVE\cite{CLIVE}$\to$KonIQ-10K$\to$KADID-10K in the~\cite{liu2021liqa}. 
We employ separate parameters to each new task by DSBN, thus avoiding the catastrophic forgetting of previously learned knowledge when learning new task (similar ideas can be found in prior works\cite{cong2020gan,lira}).  Instead of  learning new tasks in a supervised manner, we adopt unsupervised learning using our proposed self-supervised objectives. We compare our method with other five methods which learn new task in a supervised manner, including fine-tuning (FT), EWC\cite{ewc}, LWF\cite{lwf}, GFR\cite{GFR} and LIQA\cite{liu2021liqa}. We show the SROCC performance of five previously learned  datasets at the last task session in Table~\ref{tab:continual}. From the table, we can surprisingly find that although ours adopts unsupervised learning, it still shows superior performance because it does not forget the learned knowledge during the whole continual learning process. 

\vspace{-3mm}
\begin{table}[h]
\setlength{\abovecaptionskip}{0pt}
\setlength{\abovecaptionskip}{0cm}
\caption{SROCC results across previously learned datasets at the last task session.}
\scalebox{0.9}{
\begin{tabular}{ccccccc}
\hline
\multicolumn{1}{l|}{} & LIVE & CSIQ & BID & CLIVE & KonIQ-10K  \\ \hline
\multicolumn{1}{l|}{FT} & 0.819 & 0.663 & 0.398 & 0.298 & 0.573  \\
\multicolumn{1}{l|}{EWC\cite{ewc}} & 0.829 & 0.760 & 0.501 & 0.472 & 0.658  \\
\multicolumn{1}{l|}{LWF\cite{lwf}} & 0.784 & 0.721 & 0.635 & 0.497 & \textbf{0.730}  \\
\multicolumn{1}{l|}{GFR\cite{GFR}} & 0.799 & 0.702 & 0.484 & 0.301 & 0.652  \\
\multicolumn{1}{l|}{LIQA\cite{liu2021liqa}} & 0.844 & 0.705 & 0.642 & \textbf{0.572}& 0.713  \\ \hline
\begin{tabular}[c]{@{}c@{}}Ours\\ (task-aware)\end{tabular} & \textbf{0.973} & 0.750& 0.698 & 0.496 & 0.707 \\ 
\begin{tabular}[c]{@{}c@{}}Ours\\ (task-agnostic)\end{tabular} & 0.962 & \textbf{0.769} & \textbf{0.703} & 0.522 & 0.689 \\ \hline
\end{tabular}}
\label{tab:continual}
\end{table}
\vspace{-5mm}
\section{Conclusions}
In this paper, we first tackle the challenge of source-free unsupervised domain adaptation in IQA area, which is an interesting but significant topic when the source data is inavailable due to special issues. Following the observation that the rating distribution can be approximated by Gaussian, we cast the quality assessment task to a rating distribution prediction task by rating distribution construction and learning. Then to achieve a simple yet effective adaptation towards the target domain, we simply optimize the affine parameters of BN layers, which can modulate the ``source-style'' feature to the ``target-style'' feature. Considering the IQA intrinsic properties, we design three self-supervised objectives to make the predictions more consistent with the nature of human scoring nature. Extensive experiments demonstrate the effectiveness of our proposed method under cross-domain scenarios. Specially, we employ DSBN to avoid conflicts between various domains. Therefore, our method can process multiple target domains simultaneously and sequentially (continual learning setting), which is convenient and scalable.

\bibliographystyle{IEEEtran}
\bibliography{references}

\begin{thebibliography}{10}
\providecommand{\url}[1]{#1}
\csname url@samestyle\endcsname
\providecommand{\newblock}{\relax}
\providecommand{\bibinfo}[2]{#2}
\providecommand{\BIBentrySTDinterwordspacing}{\spaceskip=0pt\relax}
\providecommand{\BIBentryALTinterwordstretchfactor}{4}
\providecommand{\BIBentryALTinterwordspacing}{\spaceskip=\fontdimen2\font plus
\BIBentryALTinterwordstretchfactor\fontdimen3\font minus
  \fontdimen4\font\relax}
\providecommand{\BIBforeignlanguage}[2]{{%
\expandafter\ifx\csname l@#1\endcsname\relax
\typeout{** WARNING: IEEEtran.bst: No hyphenation pattern has been}%
\typeout{** loaded for the language `#1'. Using the pattern for}%
\typeout{** the default language instead.}%
\else
\language=\csname l@#1\endcsname
\fi
#2}}
\providecommand{\BIBdecl}{\relax}
\BIBdecl

\bibitem{uncertaintyDA}
P.~Chen, L.~Li, J.~Wu, W.~Dong, and G.~Shi, ``Unsupervised curriculum domain
  adaptation for no-reference video quality assessment,'' in \emph{Proceedings
  of the IEEE/CVF International Conference on Computer Vision}, 2021, pp.
  5178--5187.

\bibitem{rankDA}
B.~Chen, H.~Li, H.~Fan, and S.~Wang, ``No-reference screen content image
  quality assessment with unsupervised domain adaptation,'' \emph{IEEE
  Transactions on Image Processing}, vol.~30, pp. 5463--5476, 2021.

\bibitem{SHOT}
J.~Liang, D.~Hu, and J.~Feng, ``Do we really need to access the source data?
  source hypothesis transfer for unsupervised domain adaptation,'' in
  \emph{International Conference on Machine Learning}.\hskip 1em plus 0.5em
  minus 0.4em\relax PMLR, 2020, pp. 6028--6039.

\bibitem{NIMA}
H.~Talebi and P.~Milanfar, ``Nima: Neural image assessment,'' \emph{IEEE
  transactions on image processing}, vol.~27, no.~8, pp. 3998--4011, 2018.

\bibitem{zeng2017probabilistic}
H.~Zeng, L.~Zhang, and A.~C. Bovik, ``A probabilistic quality representation
  approach to deep blind image quality prediction,'' \emph{arXiv preprint
  arXiv:1708.08190}, 2017.

\bibitem{KonIQ-10k}
V.~Hosu, H.~Lin, T.~Sziranyi, and D.~Saupe, ``Koniq-10k: An ecologically valid
  database for deep learning of blind image quality assessment,'' \emph{IEEE
  Transactions on Image Processing}, vol.~29, pp. 4041--4056, 2020.

\bibitem{murray2012ava}
N.~Murray, L.~Marchesotti, and F.~Perronnin, ``Ava: A large-scale database for
  aesthetic visual analysis,'' in \emph{2012 IEEE conference on computer vision
  and pattern recognition}.\hskip 1em plus 0.5em minus 0.4em\relax IEEE, 2012,
  pp. 2408--2415.

\bibitem{ishii2021source}
M.~Ishii and M.~Sugiyama, ``Source-free domain adaptation via distributional
  alignment by matching batch normalization statistics,'' \emph{arXiv preprint
  arXiv:2101.10842}, 2021.

\bibitem{adaptiveBN}
Y.~Li, N.~Wang, J.~Shi, X.~Hou, and J.~Liu, ``Adaptive batch normalization for
  practical domain adaptation,'' \emph{Pattern Recognition}, vol.~80, pp.
  109--117, 2018.

\bibitem{autoBN}
F.~Maria~Carlucci, L.~Porzi, B.~Caputo, E.~Ricci, and S.~Rota~Bulo, ``Autodial:
  Automatic domain alignment layers,'' in \emph{Proceedings of the IEEE
  international conference on computer vision}, 2017, pp. 5067--5075.

\bibitem{DSBN}
W.-G. Chang, T.~You, S.~Seo, S.~Kwak, and B.~Han, ``Domain-specific batch
  normalization for unsupervised domain adaptation,'' in \emph{Proceedings of
  the IEEE/CVF conference on Computer Vision and Pattern Recognition}, 2019,
  pp. 7354--7362.

\bibitem{zhai2020perceptual}
G.~Zhai and X.~Min, ``Perceptual image quality assessment: a survey,''
  \emph{Science China Information Sciences}, vol.~63, no.~11, p. 211301, 2020.

\bibitem{JPEG-BIQA}
S.~Corchs, F.~Gasparini, and R.~Schettini, ``No reference image quality
  classification for jpeg-distorted images,'' \emph{Digital Signal Processing},
  vol.~30, pp. 86--100, 2014.

\bibitem{BLUR-BIQA}
X.~Wang, B.~Tian, C.~Liang, and D.~Shi, ``Blind image quality assessment for
  measuring image blur,'' in \emph{2008 Congress on Image and Signal
  Processing}, vol.~1.\hskip 1em plus 0.5em minus 0.4em\relax IEEE, 2008, pp.
  467--470.

\bibitem{SR-BIQA}
W.~Zhou, Q.~Jiang, Y.~Wang, Z.~Chen, and W.~Li, ``Blind quality assessment for
  image superresolution using deep two-stream convolutional networks,''
  \emph{Information Sciences}, vol. 528, pp. 205--218, 2020.

\bibitem{NIQE}
A.~Mittal, R.~Soundararajan, and A.~C. Bovik, ``Making a “completely blind”
  image quality analyzer,'' \emph{IEEE Signal processing letters}, vol.~20,
  no.~3, pp. 209--212, 2012.

\bibitem{liu2014no}
L.~Liu, H.~Dong, H.~Huang, and A.~C. Bovik, ``No-reference image quality
  assessment in curvelet domain,'' \emph{Signal Processing: Image
  Communication}, vol.~29, no.~4, pp. 494--505, 2014.

\bibitem{zhou2019dual}
W.~Zhou, Z.~Chen, and W.~Li, ``Dual-stream interactive networks for
  no-reference stereoscopic image quality assessment,'' \emph{IEEE Transactions
  on Image Processing}, vol.~28, no.~8, pp. 3946--3958, 2019.

\bibitem{xu2020blind}
J.~Xu, W.~Zhou, and Z.~Chen, ``Blind omnidirectional image quality assessment
  with viewport oriented graph convolutional networks,'' \emph{IEEE
  Transactions on Circuits and Systems for Video Technology}, 2020.

\bibitem{pan2018blind}
D.~Pan, P.~Shi, M.~Hou, Z.~Ying, S.~Fu, and Y.~Zhang, ``Blind predicting
  similar quality map for image quality assessment,'' in \emph{Proceedings of
  the IEEE Conference on Computer Vision and Pattern Recognition}, 2018, pp.
  6373--6382.

\bibitem{zhang2018blind}
W.~Zhang, K.~Ma, J.~Yan, D.~Deng, and Z.~Wang, ``Blind image quality assessment
  using a deep bilinear convolutional neural network,'' \emph{IEEE Transactions
  on Circuits and Systems for Video Technology}, vol.~30, no.~1, pp. 36--47,
  2018.

\bibitem{rankiqa}
X.~Liu, J.~Van De~Weijer, and A.~D. Bagdanov, ``Rankiqa: Learning from rankings
  for no-reference image quality assessment,'' in \emph{Proceedings of the IEEE
  International Conference on Computer Vision}, 2017, pp. 1040--1049.

\bibitem{zhang2021uncertainty}
W.~Zhang, K.~Ma, G.~Zhai, and X.~Yang, ``Uncertainty-aware blind image quality
  assessment in the laboratory and wild,'' \emph{IEEE Transactions on Image
  Processing}, vol.~30, pp. 3474--3486, 2021.

\bibitem{liu2021liqa}
J.~Liu, W.~Zhou, J.~Xu, X.~Li, S.~An, and Z.~Chen, ``Liqa: Lifelong blind image
  quality assessment,'' \emph{arXiv preprint arXiv:2104.14115}, 2021.

\bibitem{delange2021continual}
M.~Delange, R.~Aljundi, M.~Masana, S.~Parisot, X.~Jia, A.~Leonardis,
  G.~Slabaugh, and T.~Tuytelaars, ``A continual learning survey: Defying
  forgetting in classification tasks,'' \emph{IEEE Transactions on Pattern
  Analysis and Machine Intelligence}, 2021.

\bibitem{long2015learning}
M.~Long, Y.~Cao, J.~Wang, and M.~Jordan, ``Learning transferable features with
  deep adaptation networks,'' in \emph{International conference on machine
  learning}.\hskip 1em plus 0.5em minus 0.4em\relax PMLR, 2015, pp. 97--105.

\bibitem{zhang2015deep}
X.~Zhang, F.~X. Yu, S.-F. Chang, and S.~Wang, ``Deep transfer network:
  Unsupervised domain adaptation,'' \emph{arXiv preprint arXiv:1503.00591},
  2015.

\bibitem{gretton2006kernel}
A.~Gretton, K.~Borgwardt, M.~Rasch, B.~Sch{\"o}lkopf, and A.~Smola, ``A kernel
  method for the two-sample-problem,'' \emph{Advances in neural information
  processing systems}, vol.~19, 2006.

\bibitem{sun2017correlation}
B.~Sun, J.~Feng, and K.~Saenko, ``Correlation alignment for unsupervised domain
  adaptation,'' in \emph{Domain Adaptation in Computer Vision
  Applications}.\hskip 1em plus 0.5em minus 0.4em\relax Springer, 2017, pp.
  153--171.

\bibitem{kullback1951information}
S.~Kullback and R.~A. Leibler, ``On information and sufficiency,'' \emph{The
  annals of mathematical statistics}, vol.~22, no.~1, pp. 79--86, 1951.

\bibitem{kang2019contrastive}
G.~Kang, L.~Jiang, Y.~Yang, and A.~G. Hauptmann, ``Contrastive adaptation
  network for unsupervised domain adaptation,'' in \emph{Proceedings of the
  IEEE/CVF Conference on Computer Vision and Pattern Recognition}, 2019, pp.
  4893--4902.

\bibitem{ganin2015unsupervised}
Y.~Ganin and V.~Lempitsky, ``Unsupervised domain adaptation by
  backpropagation,'' in \emph{International conference on machine
  learning}.\hskip 1em plus 0.5em minus 0.4em\relax PMLR, 2015, pp. 1180--1189.

\bibitem{pei2018multi}
Z.~Pei, Z.~Cao, M.~Long, and J.~Wang, ``Multi-adversarial domain adaptation,''
  in \emph{Thirty-second AAAI conference on artificial intelligence}, 2018.

\bibitem{wang2020tent}
D.~Wang, E.~Shelhamer, S.~Liu, B.~Olshausen, and T.~Darrell, ``Tent: Fully
  test-time adaptation by entropy minimization,'' \emph{arXiv preprint
  arXiv:2006.10726}, 2020.

\bibitem{burkardt2014truncated}
J.~Burkardt, ``The truncated normal distribution,'' \emph{Department of
  Scientific Computing Website, Florida State University}, vol.~1, p.~35, 2014.

\bibitem{LIVE}
H.~R. Sheikh, M.~F. Sabir, and A.~C. Bovik, ``A statistical evaluation of
  recent full reference image quality assessment algorithms,'' \emph{IEEE
  Transactions on Image Processing}, vol.~15, no.~11, pp. 3440--3451, 2006.

\bibitem{CSIQ}
E.~C. Larson and D.~M. Chandler, ``Most apparent distortion: full-reference
  image quality assessment and the role of strategy,'' \emph{Journal of
  Electronic Imaging}, vol.~19, no.~1, p. 011006, 2010.

\bibitem{KADID-10K}
H.~Lin, V.~Hosu, and D.~Saupe, ``Kadid-10k: A large-scale artificially
  distorted iqa database,'' in \emph{2019 Eleventh International Conference on
  Quality of Multimedia Experience (QoMEX)}.\hskip 1em plus 0.5em minus
  0.4em\relax IEEE, 2019, pp. 1--3.

\bibitem{BID}
A.~Ciancio, E.~A. da~Silva, A.~Said, R.~Samadani, P.~Obrador \emph{et~al.},
  ``No-reference blur assessment of digital pictures based on multifeature
  classifiers,'' \emph{IEEE Transactions on image processing}, vol.~20, no.~1,
  pp. 64--75, 2010.

\bibitem{YFCC100M}
B.~Thomee, D.~A. Shamma, G.~Friedland, B.~Elizalde, K.~Ni, D.~Poland, D.~Borth,
  and L.-J. Li, ``Yfcc100m: The new data in multimedia research,''
  \emph{Communications of the ACM}, vol.~59, no.~2, pp. 64--73, 2016.

\bibitem{PIQE}
N.~Venkatanath, D.~Praneeth, M.~C. Bh, S.~S. Channappayya, and S.~S. Medasani,
  ``Blind image quality evaluation using perception based features,'' in
  \emph{2015 Twenty First National Conference on Communications (NCC)}.\hskip
  1em plus 0.5em minus 0.4em\relax IEEE, 2015, pp. 1--6.

\bibitem{BRISQUE}
A.~Mittal, A.~K. Moorthy, and A.~C. Bovik, ``No-reference image quality
  assessment in the spatial domain,'' \emph{IEEE Transactions on image
  processing}, vol.~21, no.~12, pp. 4695--4708, 2012.

\bibitem{DBCNN}
W.~Zhang, K.~Ma, J.~Yan, D.~Deng, and Z.~Wang, ``Blind image quality assessment
  using a deep bilinear convolutional neural network,'' \emph{IEEE Transactions
  on Circuits and Systems for Video Technology}, vol.~30, no.~1, pp. 36--47,
  2018.

\bibitem{HyperIQA}
S.~Su, Q.~Yan, Y.~Zhu, C.~Zhang, X.~Ge, J.~Sun, and Y.~Zhang, ``Blindly assess
  image quality in the wild guided by a self-adaptive hyper network,'' in
  \emph{Proceedings of the IEEE/CVF Conference on Computer Vision and Pattern
  Recognition}, 2020, pp. 3667--3676.

\bibitem{metaiqa}
H.~Zhu, L.~Li, J.~Wu, W.~Dong, and G.~Shi, ``Metaiqa: Deep meta-learning for
  no-reference image quality assessment,'' in \emph{Proceedings of the IEEE/CVF
  Conference on Computer Vision and Pattern Recognition}, 2020, pp.
  14\,143--14\,152.

\bibitem{adam}
D.~P. Kingma and J.~Ba, ``Adam: A method for stochastic optimization,''
  \emph{arXiv preprint arXiv:1412.6980}, 2014.

\bibitem{CLIVE}
D.~Ghadiyaram and A.~C. Bovik, ``Massive online crowdsourced study of
  subjective and objective picture quality,'' \emph{IEEE Transactions on Image
  Processing}, vol.~25, no.~1, pp. 372--387, 2015.

\bibitem{cong2020gan}
Y.~Cong, M.~Zhao, J.~Li, S.~Wang, and L.~Carin, ``Gan memory with no
  forgetting,'' \emph{Advances in Neural Information Processing Systems},
  vol.~33, pp. 16\,481--16\,494, 2020.

\bibitem{lira}
J.~Liu, J.~Lin, X.~Li, W.~Zhou, S.~Liu, and Z.~Chen, ``Lira: Lifelong image
  restoration from unknown blended distortions,'' in \emph{European Conference
  on Computer Vision}.\hskip 1em plus 0.5em minus 0.4em\relax Springer, 2020,
  pp. 616--632.

\bibitem{ewc}
J.~Kirkpatrick, R.~Pascanu, N.~Rabinowitz, J.~Veness, G.~Desjardins, A.~A.
  Rusu, K.~Milan, J.~Quan, T.~Ramalho, A.~Grabska-Barwinska \emph{et~al.},
  ``Overcoming catastrophic forgetting in neural networks,'' \emph{Proceedings
  of the National Academy of Sciences}, vol. 114, no.~13, pp. 3521--3526, 2017.

\bibitem{lwf}
Z.~Li and D.~Hoiem, ``Learning without forgetting,'' \emph{IEEE Transactions on
  Pattern Analysis and Machine Intelligence}, vol.~40, no.~12, pp. 2935--2947,
  2017.

\bibitem{GFR}
X.~Liu, C.~Wu, M.~Menta, L.~Herranz, B.~Raducanu, A.~D. Bagdanov, S.~Jui, and
  J.~v. de~Weijer, ``Generative feature replay for class-incremental
  learning,'' in \emph{Proceedings of the IEEE/CVF Conference on Computer
  Vision and Pattern Recognition Workshops}, 2020, pp. 226--227.

\end{thebibliography}

\end{document}